\pdfoutput=1
\documentclass[11pt]{article}

\usepackage{arydshln}
  
\usepackage[final]{acl}
\usepackage{booktabs}
\usepackage{times}
\usepackage{latexsym}
\usepackage{xcolor}

\usepackage[T1]{fontenc}
\usepackage[utf8]{inputenc}

\usepackage{microtype}

\usepackage{inconsolata}

\usepackage{graphicx}

\newcommand{\datasetname}{\textsc{LoraxBench}} 

\title{\datasetname: A Multitask, Multilingual Benchmark Suite for 20 Indonesian Languages}
\author{
\hspace{-1.8mm} Alham Fikri Aji$\thanks{Work done at Google.}^{\clubsuit,\diamondsuit}$ ~~ Trevor Cohn$^{\heartsuit,\diamondsuit}$  \\[3mm]
$^\clubsuit$MBZUAI ~~
$^\heartsuit$The University of Melbourne ~~
$^\diamondsuit$Google\hspace{5mm} \\
}

\usepackage{arydshln}
\begin{document}
\maketitle
\begin{abstract}

As one of the world's most populous countries, with 700 languages spoken, Indonesia is behind in terms of NLP progress. We introduce~\datasetname, a benchmark that focuses on low-resource languages of Indonesia and covers 6 diverse tasks: reading comprehension, open-domain QA, language inference, causal reasoning, translation, and cultural QA. Our dataset covers 20 languages, with the addition of two formality registers for three languages. We evaluate a diverse set of multilingual and region-focused LLMs and found that this benchmark is challenging. We note a visible discrepancy between performance in Indonesian and other languages, especially the low-resource ones. There is no clear lead when using a region-specific model as opposed to the general multilingual model. Lastly, we show that a change in register affects model performance, especially with registers not commonly found in social media, such as high-level politeness `Krama' Javanese.

\end{abstract}

% Trevor: subtle space cheats
%\addtolength{\belowcaptionskip}{-0.2ex}
%\addtolength{\abovecaptionskip}{-0.3ex}

\section{Introduction}

\begin{figure}[!ht]
\centering
\includegraphics[width=0.49\textwidth]{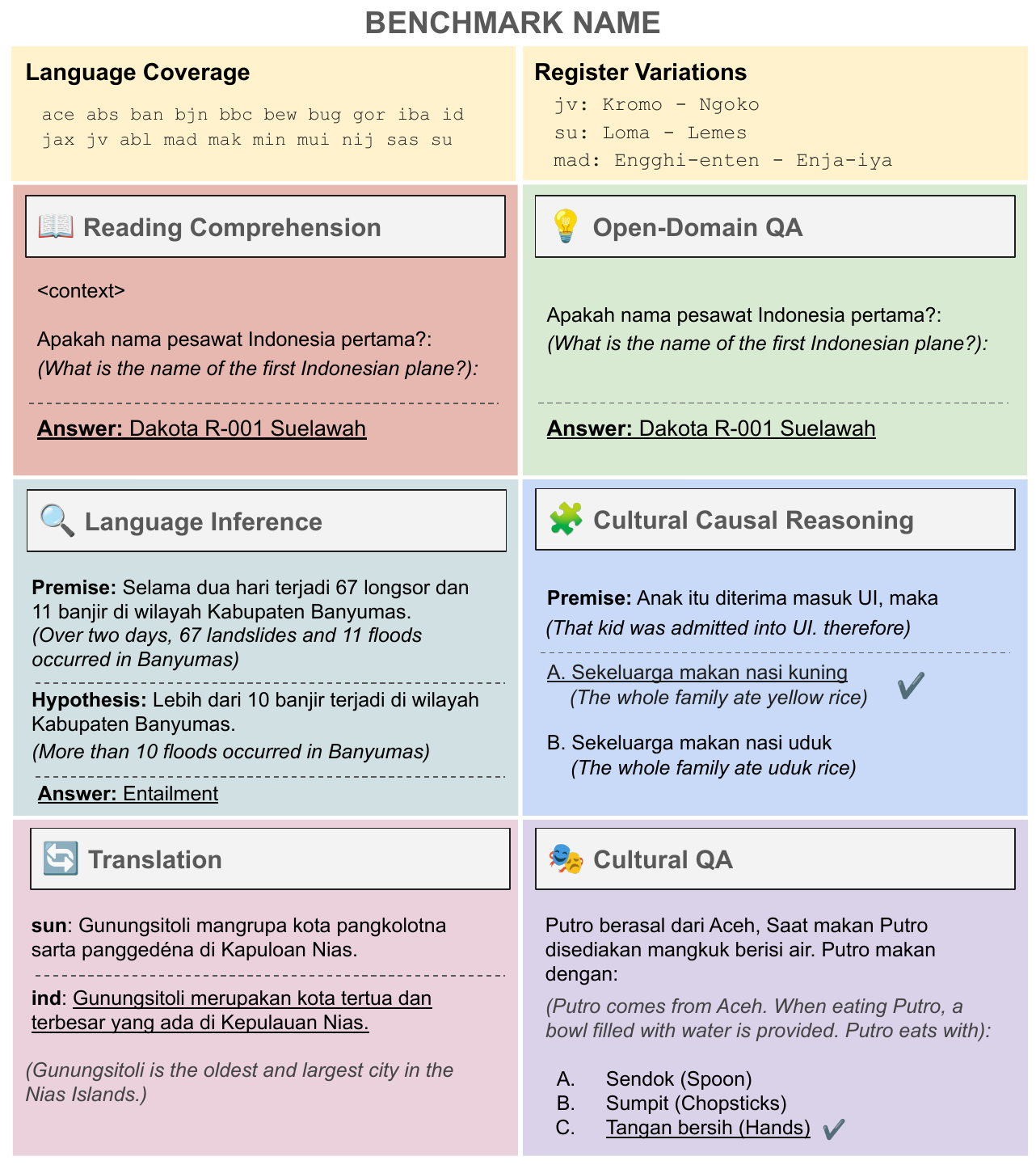}
\caption{Tasks covered in \datasetname}
\label{fig:lang_result}
\end{figure}

Indonesia is one of the world's most populous nations and one of its most linguistically diverse, being home to over 700 languages. 
Despite this, NLP research has disproportionately focused on Bahasa Indonesian and a few dominant languages, such as Javanese and Sundanese, leaving the vast majority of languages under-resourced and under-explored ~\cite{aji-etal-2022-one}.
% This includes languages like XX spoken by YY native speakers.
A major challenge for NLP in Indonesia is the lack of resources, as the shortage of data and benchmarks continues to hinder progress ~\cite{hu2020xtreme, joshi-etal-2020-state}. Moreover, even for relatively well-studied and well-resourced languages, the range of NLP tasks explored remains limited~\cite{cahyawijaya-etal-2023-nusacrowd}.

%To address this critical gap, we introduce \datasetname, a comprehensive benchmark encompassing six diverse NLP tasks: reading comprehension, machine translation, cultural reasoning, natural language inference, and cultural question answering. \datasetname~covers 20 Indonesian languages, including several low-resource languages.
% Trevor: here's my attempt as why is this important otherwise, and how our these findings generalise
%Although our dataset is specific to Indonesia and enabling NLP progress in many of its languages, the scenario we consider is common to many other parts of the world featuring rich linguistic diversity but few resources, but benefit from linguistic similarity, i.e., in terms of related languages, dialects and scripts, shared lexicon and grammar, as well as common cultural elements. 
%Progress made on improving multilingual and multicultural models on our benchmark can thus provide a pathway to accelerate progress in other important and under-served areas of the world.

To bridge this gap, we present \datasetname\footnote{\datasetname~is available at \url{https://huggingface.co/datasets/google/LoraxBench}}, a benchmark of six NLP tasks across 20 Indonesian languages: reading comprehension, machine translation, cultural reasoning, natural language inference, and cultural question answering. Our data covers many low-resource languages, including some with little to no coverage according to the comprehensive region-specific catalogue NusaCrowd~\cite{cahyawijaya-etal-2023-nusacrowd}. While focused on Indonesia, our work reflects challenges common in other linguistically diverse yet resource-scarce regions. Progress on \datasetname~can thus inform multilingual and multicultural model development globally.

Beyond multilingual capabilities, our benchmark also encompasses various registers for some languages, specifically formal and casual settings. Some of Indonesian languages, in particular, are rich in registers, with highly different use of language depending on formality of the context~\cite{rahayu2014comparison}. Unfortunately, little research has been done in this area~\cite{farhansyah2025language}. As LLMs become more integrated into daily life, such as in personal assistants, ensuring they understand and appropriately use register is vital. Our data serves as a benchmark to evaluate this capability.

We built \datasetname~by adapting existing Indonesian-language datasets through expert translation. This approach mitigates common issues with translating English datasets, especially around cultural relevance, as many concepts are widely shared across Indonesian languages but absent in Western contexts. 
%While our process may introduce some translationese, this risk is minimal given expert translators working between closely related languages. 
Moreover, the parallel nature of the data enables comparison across languages.

% Trevor: there's a nice piece in the TyDiQA paper on why avoiding English origin datasets is important. We could recycle these, although with the caveat that we introduce some manual translation biases.
% section 3.2 of https://storage.googleapis.com/tydiqa/tydiqa.pdf

We evaluate several prominent multilingual, Southeast Asian and Indonesian-focused LLMs on \datasetname, revealing significant performance disparities across languages, particularly for lower-resource languages and the more challenging polite registers, which are less represented in online data. Finally, we explore the potential of leveraging high-quality lexicons to improve model performance on specific languages.

In summary, our contributions are as follows:
\begin{itemize}
    \item We propose a new human-written benchmark for Indonesian local languages that covers 20 languages across 6 diverse tasks.
    \item For 3 languages in our benchmark we include both casual and formal registers, facilitating analysis of robustness to register.
    \item We benchmark various LLMs, from multilingual models to Indonesian-specific models, on this dataset.
\end{itemize}

\section{Related Work}

\paragraph{Benchmarks for Indonesian Languages}

Several multilingual NLP benchmarks include Indonesian, such as Flores~\cite{goyal-etal-2022-flores}, XNLI~\cite{conneau-etal-2018-xnli}, XCOPA~\cite{ponti-etal-2020-xcopa}, and Massive~\cite{fitzgerald2022massive} providing evaluation datasets for cross-lingual understanding and reasoning. However, these benchmarks typically only cover Indonesian and sometimes a small set of Indonesian local languages. Additionally, their English-centric data construction often results in content that is not contextually relevant.

To address these limitations, dedicated efforts have been made to develop benchmarks with a stronger focus on Indonesian-specific content. Examples include IndoNLU~\cite{wilie-etal-2020-indonlu}, IndoNLI~\cite{mahendra-etal-2021-indonli}, IndoMMLU~\cite{koto-etal-2023-indommlu}, 
and COPAL-ID~\cite{wibowo-etal-2024-copal}. Other benchmarks, such as NusaWrites~\cite{cahyawijaya-etal-2023-nusawrites} and NusaX~\cite{winata-etal-2023-nusax}, have been designed to evaluate regional languages, typically covering low-resource Indonesian languages. Our work improves in this direction by providing benchmark with Indonesian-relevant content that covers more languages and tasks.

\paragraph{Benchmarks for Low-Resource Languages}

Beyond benchmarks for Indonesian languages, we also see recent progress in benchmarks for other languages, especially those that are underexplored. Efforts such as MasakhaNER~\cite{adelani-etal-2021-masakhaner, adelani-etal-2022-masakhaner} and MasakhaNews~\cite{adelani-etal-2023-masakhanews} are enriching datasets for African languages, while initiatives for Indic languages, such as IndicNLP Suite~\cite{kakwani-etal-2020-indicnlpsuite}, are driving similar advancements in the South Asian context. These efforts help to address the data gap for low-resource languages, facilitating more inclusive and robust language models across diverse communities. % We also see largely multilingual benchmarks that covers languages beyond subregion, such as 

We also see several massively multilingual benchmarks that cover languages across the globe. MASSIVE~\cite{fitzgerald2022massive} is an intent-classification task for 60 languages. Belebele ~\cite{bandarkar2023belebele} is a large-scale reading comprehension benchmark covering 122 languages and language variants. Flores~\cite{goyal2022flores} is a machine translation benchmark covering 200 languages. INCLUDE~\cite{romanou2024include} and Global-MMLU~\cite{singh2024globalmmluunderstandingaddressing} are exam-like benchmarks for more than 40 languages. Despite their broad coverage, only a small fraction of Indonesian languages are included, typically Indonesian, Javanese, and Sundanese. Not only that, these benchmarks often were translated from English, resulting in context bias that might not fully capture cultural nuances in Indonesia~\cite{mihalcea2024ai}. Our proposed benchmark covers more languages that are not typically included in massively multilingual benchmarks.

% \textbf{Trevor: We need to mention things like Flores, Belebele and the like. We could also assess how much of these overlap with our languages of focus, which would be handy. We could even `include' slices of these datasets in our benchmark, although I think the overlapping languages may be too few to paint a clear picture.}

\section{\datasetname}

\subsection{Language of Focus}

\begin{table}[h]
\centering
\small
\begin{tabular}{@{}l@{}rl@{}}
\hline
\textbf{Language} & \textbf{Speakers} & \textbf{Spoken in} \\
\hline
Acehnese (ace) & 3.7 M & Aceh \\
Ambonese Malay (abs) & 0.2 M & Ambon \\
Balinese (ban) & 4.8 M & Bali \\
Banjar (bjn) & 4.0 M & South Sulawesi \\
Batak Toba (bbc) & 2.5 M & North Sumatra \\
Betawi (bew) & 5.6 M & Jakarta \\
Buginese (bug) & 4.3 M & South Sulawesi \\
Gorontalo (gor) & 1.1 M & Gorontalo \\
Iban (iba) & 0.8 M &  West Kalimantan \\
Jambi Malay (jax) & 1.0 M & Jambi  \\
Javanese (jv) & 91.0 M & East/Central Java \\
Lampung Nyo (abl) & 1.5 M & Lampung \\
Madurese (mad) & 17.0 M & East Java \\
Makasar (mak) & 1.9 M & Makasar \\
Minangkabau (min) & 8.0 M & West Sumatra \\
Musi (mui) & 3.1 M & South Sumatra \\
Ngaju (nij) & 0.9 M & Central Kalimantan \\
Sasak (sas) & 2.6 M & West Nusa Tenggara \\
Sundanese (su) & 32.0 M & West Java \\
\hline
\end{tabular}
\caption{Statistics of the languages of focus in \datasetname, based on LinguaMeta~\cite{ritchie-etal-2024-linguameta-unified}}
\label{tab:lang-coverage}
\end{table}

This work focuses on Indonesian and 19 Indonesian local languages, representing a diverse range of population sizes and geographical regions, as detailed in Table~\ref{tab:lang-coverage}.  Several of these languages have not previously been included in public downstream NLP tasks, as evidenced by their absence in comprehensive catalogs like SEACrowd~\cite{lovenia-etal-2024-seacrowd} and NusaCrowd~\cite{cahyawijaya-etal-2023-nusacrowd}. Specifically, excluding unlabeled corpora, word lists, and lexicons, \texttt{iba}, \texttt{jax}, and \texttt{sas} were absent from the SEACrowd text data catalogue, while \texttt{bbc}, \texttt{bew}, \texttt{gor}, and \texttt{mui} only have translation or sentiment analysis downstream tasks.

One  key challenge in Indonesian NLP is the diversity of registers across languages~\cite{aji-etal-2022-one}.  Usage often varies significantly between formal and informal settings.  Existing datasets frequently overlook this nuance, typically focusing solely on the casual register. To address this gap, our dataset includes an additional formal register variation for Sundanese, Javanese, and Madurese. In total, the data encompasses Indonesian, 19 local languages and 3 additional registers, resulting in 23 distinct subsets.

\subsection{Formal and Casual Registers}

Prior NLP research on these languages often overlooks the granularity and diversity of local language registers in Indonesia~\cite{farhansyah2025language}. Therefore, for Javanese, Sundanese, and Madurese, we gather data across two different registers: one more formal and one more casual. Each of these languages has distinct levels of politeness used in different conversational settings, whether with peers or in more refined, formal situations.\footnote{Some Indonesian languages have further distinct registers (Sundanese has 6), however few people are fluent in all registers; pragmatically our selection of two registers cover most common usage.}

In this work, we select two registers for each language. Specifically, for Javanese, we use \textbf{Krama} as the formal register and \textbf{Ngoko} as the casual register. Similarly, for Sundanese, we use \textbf{Lemes} (formal) and \textbf{Loma} (casual), while for Madurese, we use \textbf{Engghi Ethen} (formal) and \textbf{Enja'Iya} (casual).

In all cases, the formal registers are typically used when conversing with individuals of higher status, such as parents, bosses, or, in some cases, strangers, whereas the casual registers are used with peers and friends~\cite{hadiwijaya2017kesantunan}. These registers differ significantly, particularly in vocabulary. For example, `me' is \textit{kula} in formal Javanese but \textit{aku} in casual settings. Similarly, `want' is \textit{badé} in formal Sundanese but \textit{arék} in casual contexts. Using an incorrect register may come across as impolite or awkward.

\subsection{Task Coverage}

\datasetname~covers 6 different tasks. We use Indonesian data source, to ensure contextual relevance of the dataset after its translation. % to local languages.

\paragraph{Reading Comprehension} We adopt the Indonesian set from TyDi QA~\cite{clark2020tydi} for reading comprehension, which is based on Indonesian Wikipedia with human-written questions. Specifically, we take the secondary task of Tydi-QA, where it is given a passage and a question, and the answer is the span from the text. Different from the rest of data that we use, Tydi-QA consisted of training and test split, therefore we translate the test set alongside 100 sampled training instances, which can be used as a small training split.

\paragraph{Open-Domain QA} By removing the context from our reading comprehension task, we can repurpose it into an open-domain QA task, where the model must rely on its internal knowledge to answer the question.

\paragraph{Natural Language Inference} For NLI, we translate from IndoNLI~\cite{mahendra-etal-2021-indonli}. Specifically, IndoNLI consisted of crowd-written and expert-written instances, where the latter is more challenging and of higher quality. The expert-written data covers complex tasks such as temporal and numerical reasoning, but only covers the test split. Therefore, we translate the expert-written test split of IndoNLI, specifically for the single-sentence subset. 

\paragraph{Machine Translation} As IndoNLI premises were collected from Indonesian sites and local websites, covering various domains, we can also re-purpose the translation of IndoNLI premises as our machine translation benchmark. We evaluate the into Indonesian direction, from each source language.
% Trevor: please confirm the last bit about the direction; and do we need to defend this choice (IIRC that it's better to eval the reverse task as that done by manual translators, at least this means we get better data coverage)

\paragraph{Causal Reasoning} We take COPAL-ID~\cite{wibowo-etal-2024-copal} as our causal reasoning data. COPAL is similar to COPA~\cite{wang2019superglue}, where we give a premise, and the model must chose among the most likely cause/effect. However, unlike COPA, COPAL is carefully handcrafted and contains cultural and local nuances, therefore presenting additional challenges.

\paragraph{Cultural QA} Lastly, we translate IndoCulture~\cite{koto-etal-2024-indoculture} for cultural QA. We select the non-province specific set, to avoid questions that are specific to a particular province and may not be relevant after translation.

% \textbf{A big table (full width) with examples illustrating each task would be handy. Each could feature a different language or register.}

\subsection{Data Creation}

\paragraph{Data Clean-up}
For COPAL and IndoCulture, we observed that the data required filtering and cleanup. The COPAL data is heavily Jakartan-centric in terms of cultural reasoning. To avoid overly specific cultural understanding, especially when translating the questions into other languages, we manually remove such data. Filtering of entries was done by a native speaker who has lived in Jakarta, Indonesia. After filtering, we ended up with 365 instances to be translated. We will release the filtered COPAL version alongside this work.

\begin{table}
\centering
\begin{tabular}{lc}
\toprule
\textbf{Category} & \textbf{Count} \\
\midrule
Removed & 61 \\
Fixed Typos & 12 \\
Improved Distractors & 13 \\
\textbf{Final Data Count} & 510 \\
\bottomrule
\end{tabular}
\caption{IndoCulture Cleanup Statistics}
\label{tab:indoculture_cleanup}
\end{table}

The IndoCulture questions also required cleanup. Specifically, we observed that the text quality is sometimes poor, as a result of crowdsourced data collection. Therefore, first, we fix writing errors and typos. We also note that some distractors in the multiple-choice questions are trivially incorrect. For example, having a "rocket" as a mode of transportation to the local market is obviously incorrect. In this case, we change the distractors into more believable options. 

Lastly, we note that some questions are repetitive. Similarly, some questions are arguably obvious, as they do not really ask for culture-specific information but rather focus on good manners, often paired with an obviously incorrect answer. For example, ``Your close family member has just died; you must,'' with the correct answer being ``to help the family'' and the incorrect one being ``to insult them.'' We manually remove such questions. Statistics of the cleaned IndoCulture dataset are in Table~\ref{tab:indoculture_cleanup}.

For OpenQA, we note that removing the passage might render some questions unanswerable. Therefore, we manually validate all questions to determine whether they are still answerable without context. We identified only six such questions. These include questions whose answers could change over time, such as the location of an office or the youngest chess grandmaster, or that become ambiguous without context, such as a popular travel destination in a given area. We then remove these questions.

\paragraph{Data Translation}

For all tasks, we translate the Indonesian instances into the corresponding languages via professional translation. Annotators are native speakers. 
% Translation instructions can be found in the Appendix. 
Validation is performed through human review, where each entry is validated by another native speaker. On top of that, we employ automated methods to assist human validation. Specifically, our automatic validation detects potentially incorrect translations by identifying anomalies in translation length and numerical inconsistencies. All issues were flagged for validation and error cases re-translated. Our team held several meetings with the annotators to discuss annotation guideline, concerns and address inconsistencies, until we were satisfied with the data. The  instructions to annotators are provided in Appendix~\ref{appendix:annotation}.

\subsection{Resulting Data}

\begin{table}
\small
\centering
\begin{tabular}{@{}l@{ ~ }c@{ ~ }c@{ ~ }c@{}}
\toprule
\textbf{Category} & \textbf{Source} & \textbf{\# Lang /} & \textbf{Total} \\
 &  & \textbf{Register} & \textbf{Examples} \\

\midrule
Causal Reasoning & COPAL & 23 & 8395 \\
Language Inference & IndoNLI & 23 & 33258\\
Cultural QA & IndoCulture & 23 & 11730 \\
Reading Comprehension & Tydi-QA & 23 & 12972 \\
Open-Domain QA & Tydi-QA & 23 & 12834 \\
Translation & IndoNLI & 22 & 5522 \\
\midrule
\textbf{Total} & & & 84711 \\
\bottomrule
\end{tabular}
\caption{\datasetname~test size}
\label{tab:data-stat}
\end{table}

We summarize \datasetname~in Table~\ref{tab:data-stat}, which consists of a total of 84,895 data points across 6 diverse tasks and 23 languages and registers. Our data will be made publicly available, with an unrestrictive licence.

\begin{table*}[]
    \centering
    \small
    \begin{tabular}{ll}
        \toprule
         & Example \\
        \midrule
        Indonesian & Setelah pertempuran melawan Romawi, Muawiyah dan tentaranya menang.
        \\
        English & After a battle against the Romans, Muawiyah and his soldiers were victorious. \\
        \midrule
        Krama Javanese & Sasampunipun perang nglawan tiyang-tiyang Romawi, Muawiyah lan prajuritipun kasil menang. \\
        Ngoko Javanese & 
Sawise tarung nglawan wong-wong Romawi, Muawiyah karo prajurite iso menang. \\
        \midrule
        Lemes Sundanese & Saatos tarung ngalawan jalmi-jalmi Romawi, Muawiyah sarta soldadu na junun kenging. \\
        Loma Sundanese & Sanggeus perang ngalawan Romawi, Muawiyah jeung pasukan meunang.\\
        \midrule
        Engghi Ethen Madurese & Saampon atokar bhleben reng-oreng Romawi, Muawiyah ben prajuritnah ahasel menang. \\
        Enja'Iya Madurese & Semarena tarong ngelaben oreng-oreng Romawi, Muawiyah ben prajuritta hasel menang.\\
        \bottomrule
    \end{tabular}
    \caption{Examples of polite and casual register differences. The sentences above are parallel.}
% Trevor: add the Indonesian source sentence in the caption. These are all the same sentence, ja?
    \label{tab:formal-example}
\end{table*}

\begin{table}[h!]
    \centering
    \small
    \begin{tabular}{@{}r@{ ~ }ccc@{}}
    \toprule
         & \multicolumn{1}{c}{Javanese} & \multicolumn{1}{c}{Sundanese} & \multicolumn{1}{c}{Madurese} \\
    \midrule 
    \multicolumn{3}{c}{\textbf{Vocabulary Overlap}} \\
         $|V_F|$ & 3621 & 3559 & 4273\\
         $|V_C|$ & 3756 & 3590 & 4592\\
         $|V_F \cap V_C|$ & 2130 & 2540 & 2142 \\
    \midrule
    % \multicolumn{3}{c}{\textbf{Vocabulary Differences Examples}} \\
    % I\hspace{15}$F$ & Kula & Urang & \\
    % $C$ & Aku & Abdi \\
    % \hdashline You\hspace{15}$F$ & Sampéyan & Anjeun & \\         
    % $C$ & kowé & Maneh \\
    % \hdashline He/she\hspace{15}$F$ & Piyambakipun & Manehna & \\
    % $C$ & Dhèwèké & Mantenna\\
     
    % \hdashline Drink\hspace{15}$F$ & Ngunjuk & Leueut & \\
    % (verb)\hspace{15}$C$ & Ngombé & Nginum\\
    % \hdashline No\hspace{15}$F$ & Mboten & Henteu & \\
    % $C$ & Ora & Enteu & \\
     
        % Top $V_F - V_C$ & ingkang:467 &  langkung:76 & panekah:263 \\
       %  & dipun:283 & anjeunna 50 & lok:169\\
        % & punika:281 & seueur:43 & ebektoh:86\\
        % Top $V_C - V_F$ & karo:295 & téh:133 & ndek:115\\
        % & seng:265 & loba:64 & jiah:96 \\
        % & ning:254 & boga:60 & ajiyeh:59 \\
    % \midrule
    \multicolumn{3}{c}{\textbf{Sentences Differences}} \\
    BLEU($F$, $C$) & 8.03 & 13.0 & 7.3 \\
    Jaccard($F$, $C$) & 0.11 & 0.16 & 0.10 \\
    \bottomrule
    \end{tabular}
    \caption{Formal and Casual differences on lexical level. $F$ and $C$ denotes formal and casual data, whereas $V_{F|C}$ denotes their respective vocabulary.}
    \label{tab:formal-vs-casual}
\end{table}

To further understand the differences between these registers, we analyzed word overlap, as detailed in Table~\ref{tab:formal-vs-casual}. The results indicate substantial vocabulary variation between the formal and casual registers, with Sundanese exhibiting the highest degree of similarity. These registers differs in some of commonly use word such as pronouns or common verbs. Examining the most frequent words unique to each register reveals numerous function words, such as particles (e.g., téh in Sundanese) and prepositions (e.g., karo `with', ning `in'). This suggests significant divergence in word usage, even for common lexical items. Beyond the vocabulary level, we confirm that their sentence-level differences are noticeable, as shown by their sentence-level BLEU$_4$ or Jaccard similarity, or as illustrated by the example in Table~\ref{tab:formal-example}.

\section{Experiment Setup}

\subsection{Models}

We benchmark several language models in a zero-shot manner across all of our tasks. We explore leading multilingual foundation models such as BLOOMZ~\cite{muennighoff-etal-2023-crosslingual}, Gemma 2~\cite{team2024gemma} and 3~\cite{team2025gemma}, Gemini 1.5~\cite{team2024gemini} and 2.5~\cite{comanici2025gemini}, Aya-23~\cite{aryabumi2024aya}, and QWEN-2.5~\cite{yang2025qwen2}. Additionally, we evaluate models specifically designed for the Southeast Asian region, such as Sailor~\cite{dou-etal-2024-sailor}, SEA-LION, and SeaLLM~\cite{damonlp2024seallm3}. Lastly, we examine Indonesian-specific models, including Cendol~\cite{cahyawijaya-etal-2024-cendol} and Sahabat-AI.\footnote{\url{https://sahabat-ai.com/}} Specific model checkpoints used are listed in Appendix~\ref{appendix:models}. 

% For each task, we employ several baseline prompts and report the maximum performance. Specifically, for classification tasks (NLI, COPAL, CulturalQA), we experiment with both free-text generation and a log-probability-based method to predict the label.

\begin{figure*}[ht]
    \centering
    \begin{minipage}{0.49\textwidth}
        \centering
        \includegraphics[width=\textwidth]{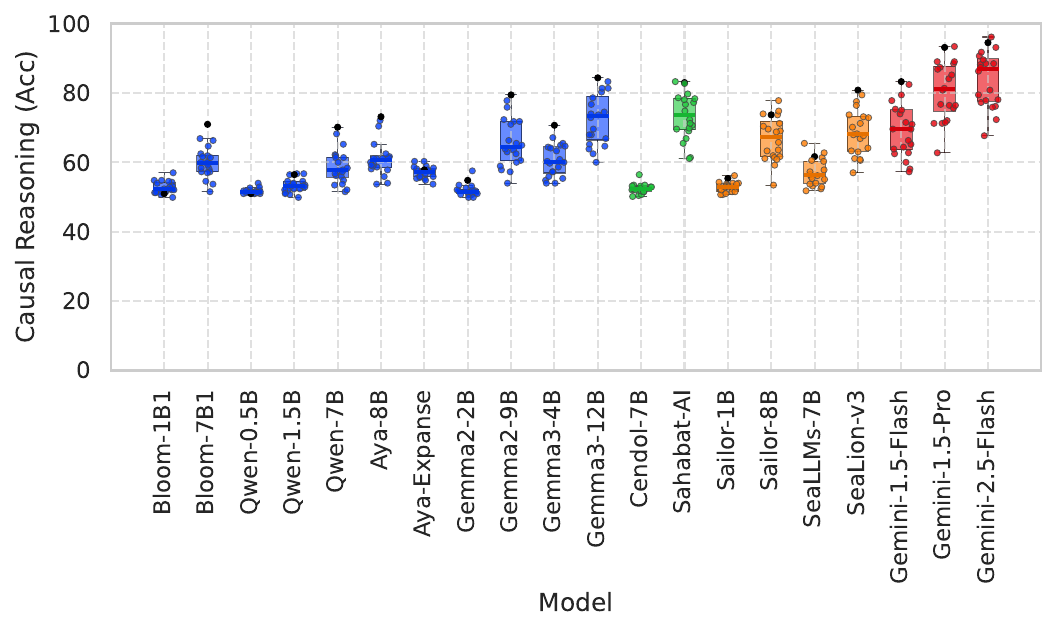}
        
    \end{minipage}
    \hfill
    \begin{minipage}{0.49\textwidth}
        \centering
        \includegraphics[width=\textwidth]{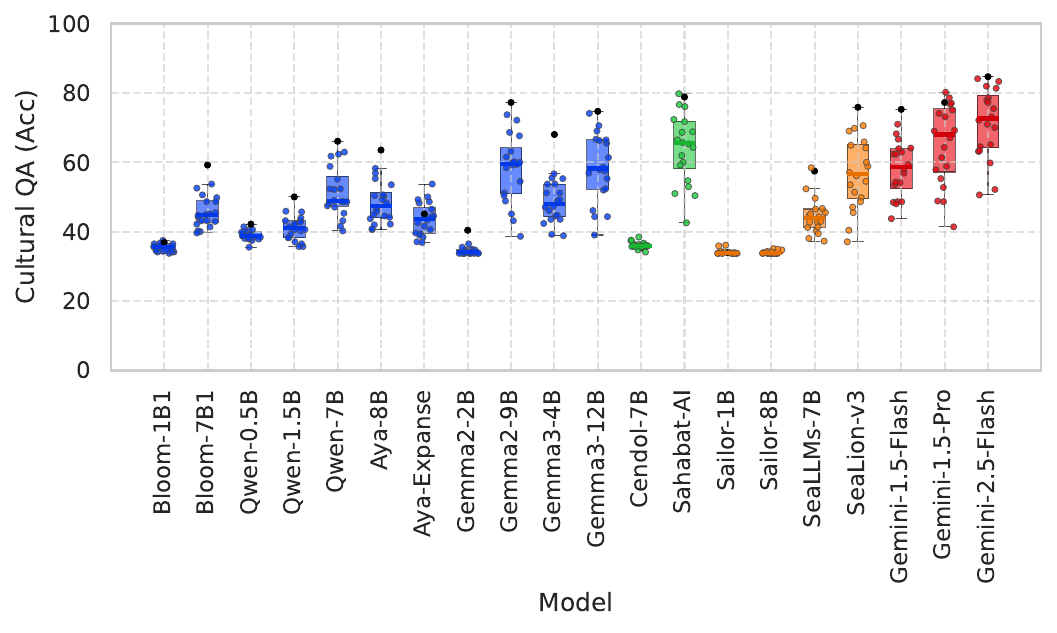}
    \end{minipage}
    
    \vspace{0.1cm}
    
    \begin{minipage}{0.49\textwidth}
        \centering
        \includegraphics[width=\textwidth]{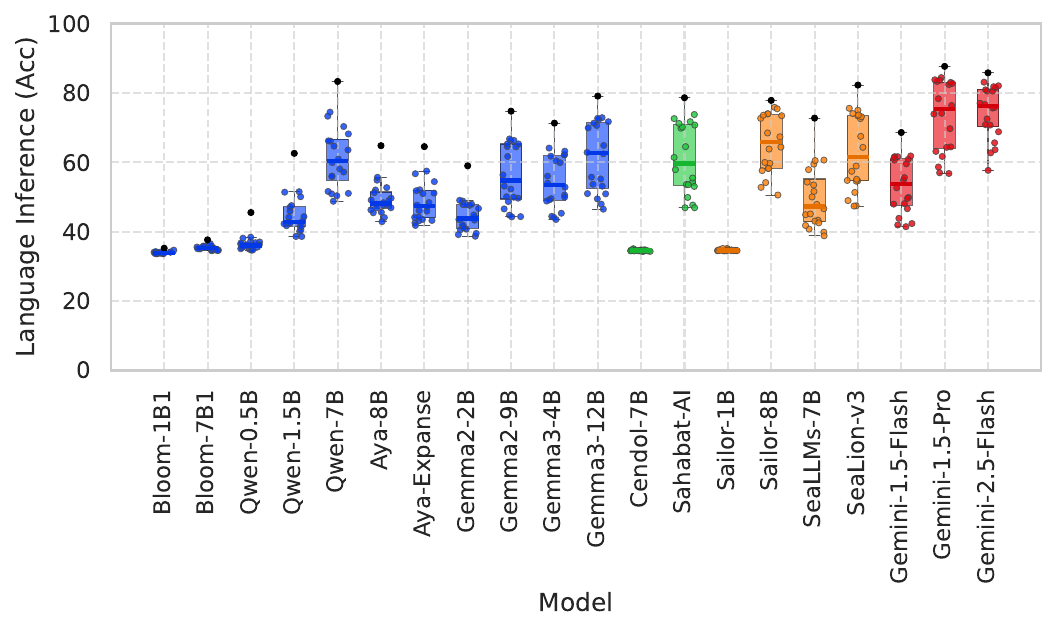}
    \end{minipage}
    \hfill
    \begin{minipage}{0.49\textwidth}
        \centering
        \includegraphics[width=\textwidth]{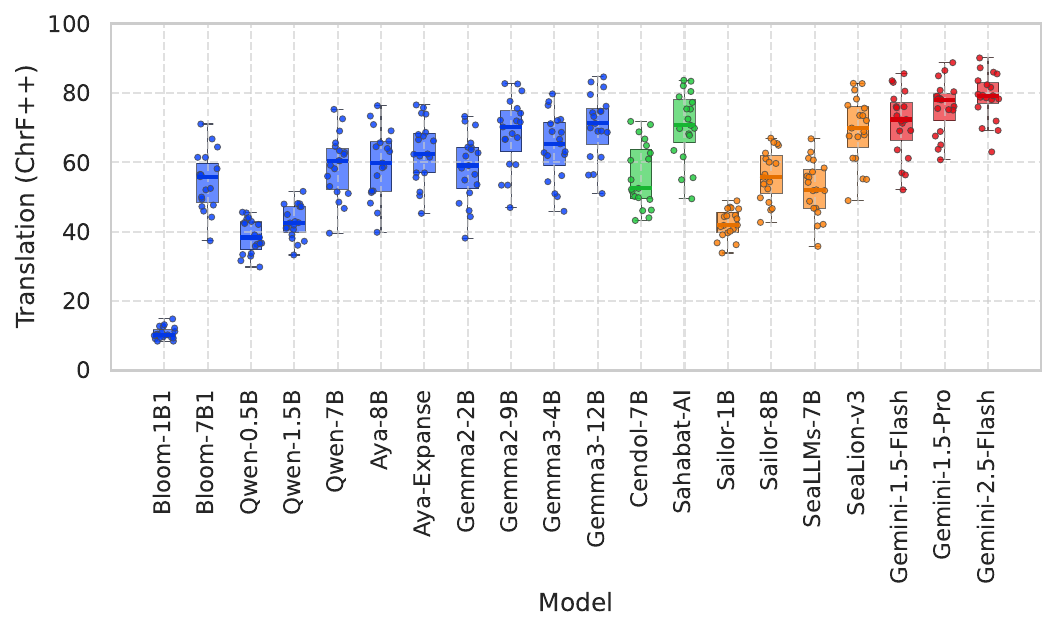}
    \end{minipage}
    
    \vspace{0.1cm}
    
    \begin{minipage}{0.49\textwidth}
        \centering
        \includegraphics[width=\textwidth]{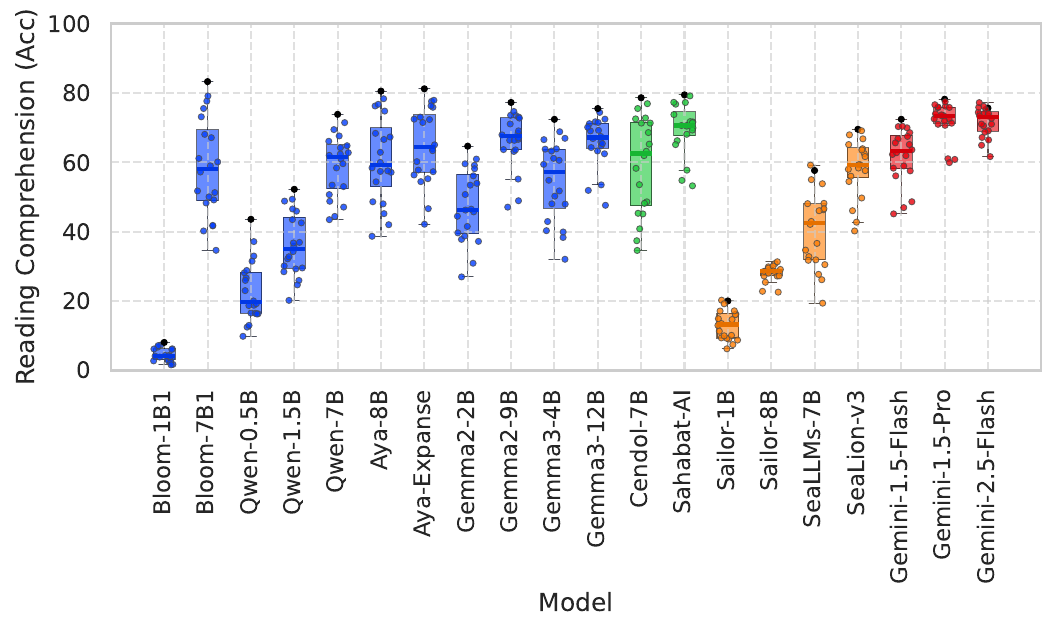}
    \end{minipage}
    \hfill
    \begin{minipage}{0.49\textwidth}
        \centering
        \includegraphics[width=\textwidth]{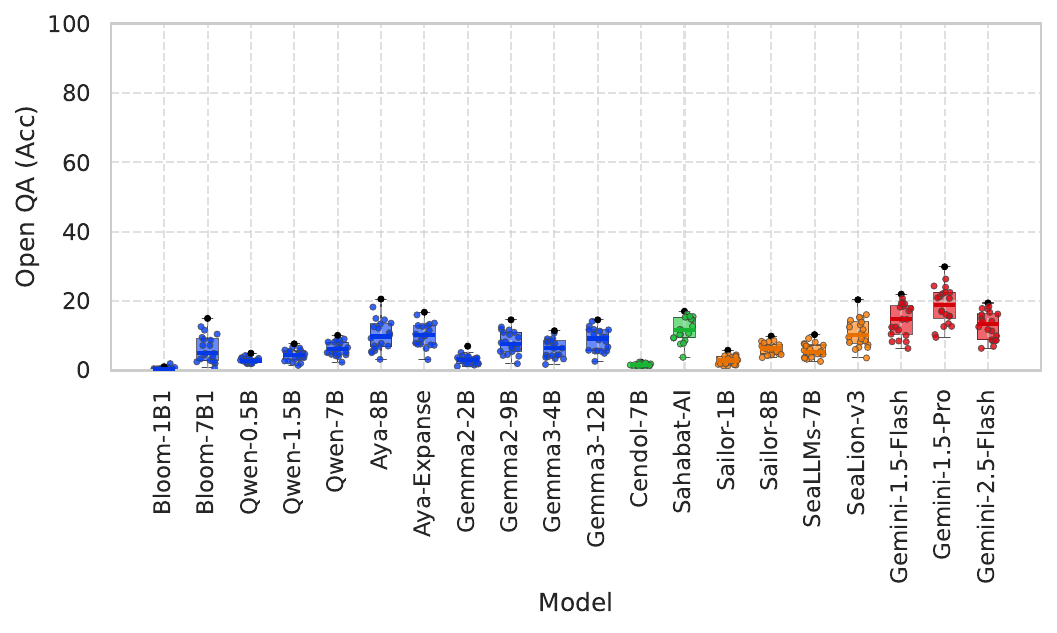}
    \end{minipage}
    
    \caption{Results across different models. Black dot indicates performance in Indonesian.}
    \label{fig:res_model}
\end{figure*}

\subsection{Prompt Design}
% \textbf{Trevor: An Appendix with all the prompts listed would be useful.}
As a baseline, we employ a standard prompt that directly asks the models for answers. However, we explore prompt strategy variations that leverage linguistic similarities with Indonesian. In addition, we experiment with extracting answers directly from the generated text and selecting the answer with the highest log-probability, though the latter is only applicable to classification tasks for publicly available models. We report the maximum performance across all prompts used.

\paragraph{Language-Informed} The local languages of Indonesia typically share similar grammar with Indonesian. Moreover, they share some vocabulary. We can exploit this information by explicitly informing the LLM via prompts.

% \paragraph{Few-shot Translation} To further familiarize the model with the language, we utilize the training set of TyDi-QA as a few-shot translation example.

\paragraph{Lexicon-Guided} Lastly, we introduce a method to enhance the model's understanding of the language by incorporating lexicon information. Specifically, we utilize the Gatitos lexicon~\cite{jones-etal-2023-gatitos}, a high-quality, human-crafted lexicon between several languages and English. For inputs written in local languages, we retrieve all available word translations and provide them as additional prompt information in the format \texttt{<word> → <translation1>, <translation2>, ...}, with each line representing a word found in the lexicon.  Gatitos does not cover all our languages, therefore we apply the approach to those supported languages.

\paragraph{Log-probability-based and Cloze} Some of our tasks are multiple-choice (Language Inference, Cultural QA, Causal Reasoning). Therefore, in addition to free-text generation, we also select the label based on the most likely choice generated by the model, using log probabilities. Moreover, Cultural QA and Causal Reasoning can be framed as cloze tasks, by simply concatenating the context sentence with the possible answers and selecting the one with the highest probability (e.g., `it is raining' therefore `it is wet outside'). For these two tasks, we additionally incorporate cloze-based prompting with log probabilities. Since this approach requires access to the model’s probability outputs, it is only applicable to open models.

\paragraph{Few-Shot Prompting} Specifically for reading comprehension using Tydi-QA, we also have a small amount of training data that can be used for few-shot prompting. Therefore, we additionally explore few-shot prompting for this particular task.
%\textbf{Trevor: How might we improve LRL performance? We could do continued pre-training of Gemma with the NTL dataset (&/or Madlad), and/or add automatic translations from xx-id and id-xx. Maybe we don't need to do this for a benchmark paper, but I do wonder.}

\section{Results}

% \textbf{Trevor: This figure is pretty busy. Perhaps aggregating non-id languages in one column (or IQR/whisker style chart) would be simpler, and have the 5 tasks side-by-side. We could measure the gap vs id performance, rather than absolute accuracy, too. Hmm, would need to try some visualisations to see.}

Figure~\ref{fig:res_model} lists the model's performance across different task, for each languages. We select the best-performing prompt for each setting, and we report the accuracy for all tasks except for translation in which we use ChrF++. The Open QA system is considered correct if the answer exists in the generated response. We do note that exact-match approach might be too strict, so in addition we also contrast it with LLM-as-a-judge evaluation in Appendix~\ref{appendix:openqa}.

\subsection{Result Across Models}

%\begin{figure}[ht!]
%\centering
%\includegraphics[width=0.45\textwidth]{latex/figures/aggregate_Model.pdf}
%\caption{Aggregated performance across model}
%\label{fig:model_result}
%\end{figure}

Generally, larger models outperform their smaller counterparts within the same model family (e.g., Qwen). Beyond this, Sahabat-AI, an Indonesian-focused model based on Gemma, has a slight edge over the other models. Interestingly, SeaLion-v3, also based on Gemma, does not show the same improvement, highlighting the crucial role of continual fine-tuning design.

We also observe a lack of consistency across tasks. One model may excel in a particular task, while another may perform better in a different task. This inconsistency is evident even within the Gemma family and its two derivatives, Sahabat-AI and SeaLion. Sahabat-AI notably improves Gemma's performance on the NLI and causal reasoning tasks, particularly on the causal reasoning side. In contrast, SeaLion shows a drop in QA performance. We can also see a similar pattern with Aya and Qwen, where Aya is stronger than Qwen in some sets but weaker in others.

Commercial models like Gemini exhibit solid performance, but the gap is not as significant when compared to publicly available models.

\subsection{Result Across Tasks}

\begin{figure*}[ht]
    \centering
    \begin{minipage}{0.49\textwidth}
        \centering
        \includegraphics[width=\textwidth]{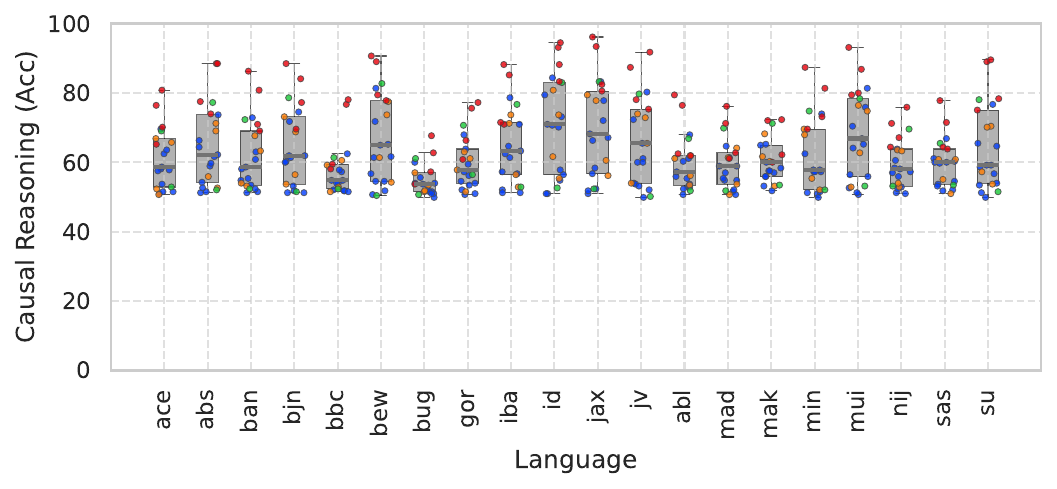}
        
    \end{minipage}
    \hfill
    \begin{minipage}{0.49\textwidth}
        \centering
        \includegraphics[width=\textwidth]{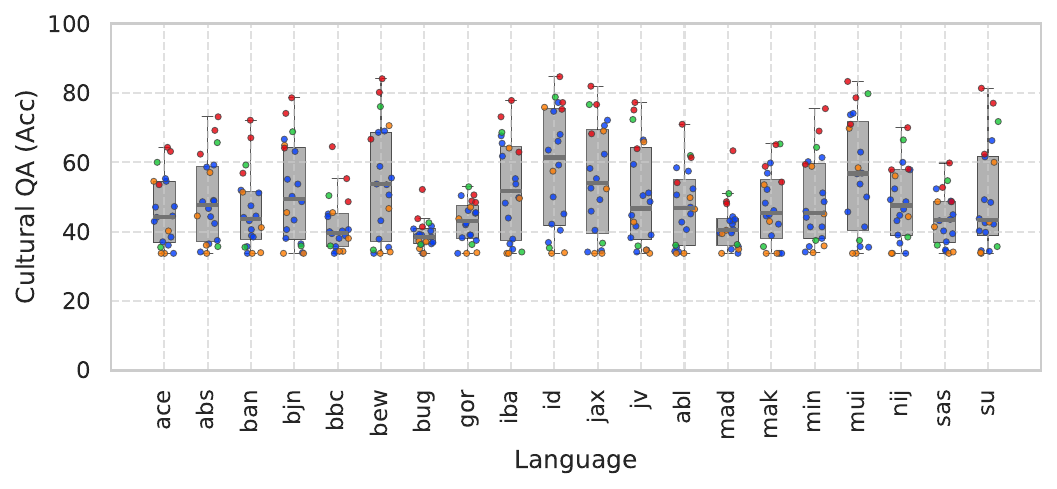}
    \end{minipage}
    
    \vspace{0.1cm}
    
    \begin{minipage}{0.49\textwidth}
        \centering
        \includegraphics[width=\textwidth]{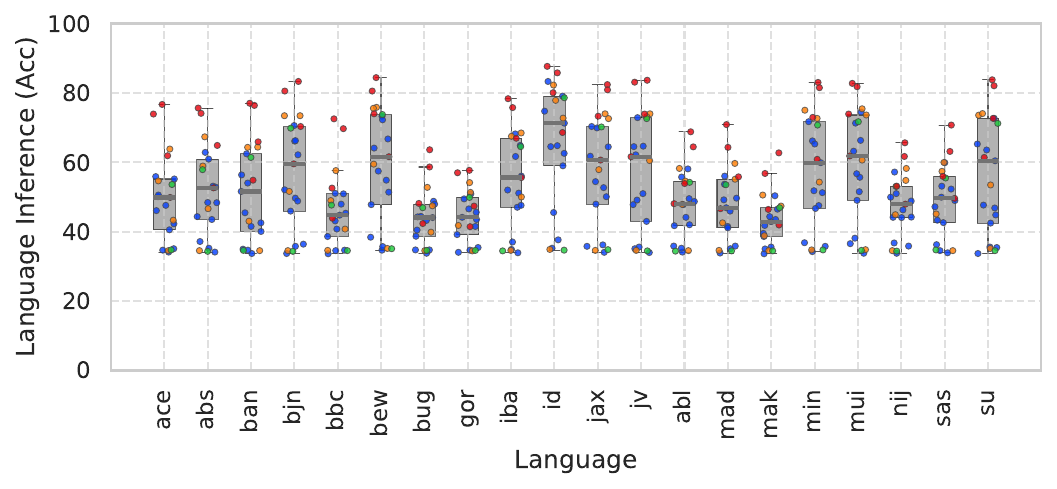}
    \end{minipage}
    \hfill
    \begin{minipage}{0.49\textwidth}
        \centering
        \includegraphics[width=\textwidth]{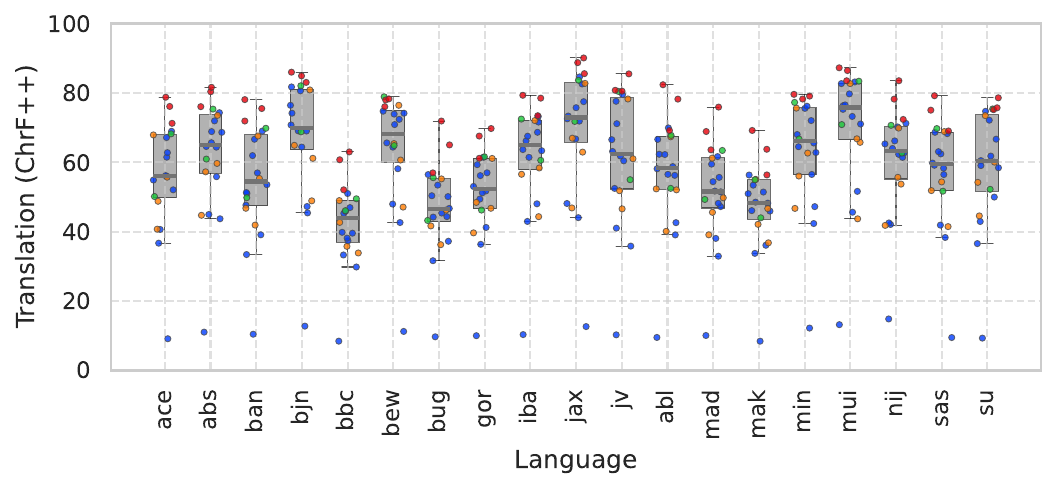}
    \end{minipage}
    
    \vspace{0.1cm}
    
    \begin{minipage}{0.49\textwidth}
        \centering
        \includegraphics[width=\textwidth]{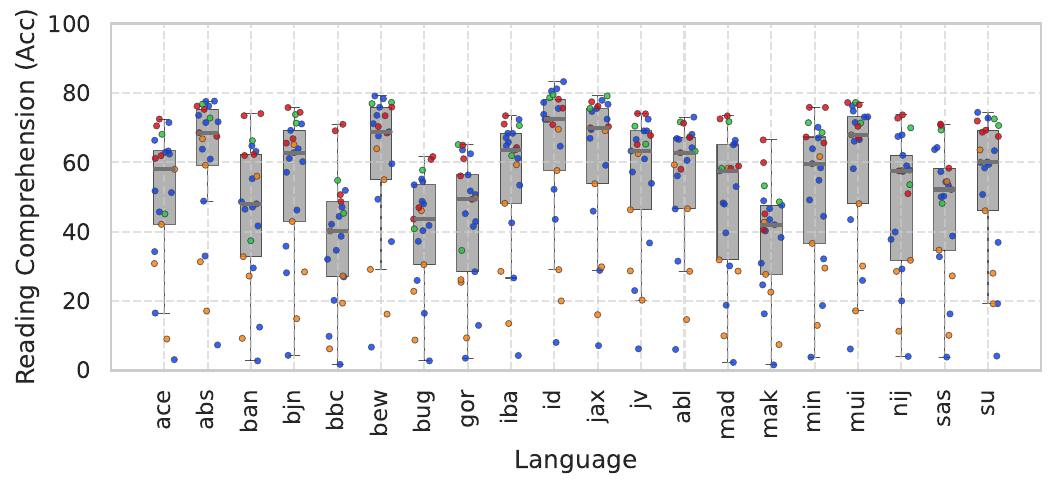}
    \end{minipage}
    \hfill
    \begin{minipage}{0.49\textwidth}
        \centering
        \includegraphics[width=\textwidth]{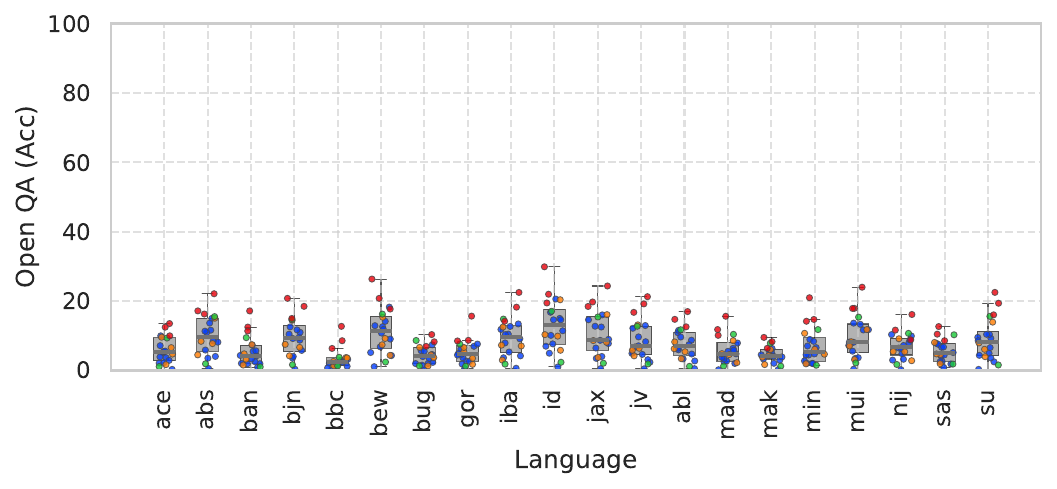}
    \end{minipage}
    
    \caption{Results across different languages. Each dot represents a different model, with colors corresponding to \textcolor{blue}{multilingual open models}, \textcolor{red}{commercial models}, \textcolor{green}{SEA-specific models}, or \textcolor{orange}{Indonesian-specific models}.
}
    \label{fig:res_lang}
\end{figure*}

\paragraph{Causal Reasoning} The culturally relevant causal reasoning task is based on COPAL-ID, where the model is required not only to reason causally but also to understand cultural and local nuances. The original Indonesian data is already challenging, with most models achieving close to random guessing of around 50\%, while humans can easily achieve 95\%, according to their report. Only a handful of models manage to achieve reasonably high scores.

Many general multilingual models perform poorly, including more recent and larger variants such as Aya and Qwen. The exception is Gemma-9B, which generally outperforms the other multilingual models. This is expected, as this data requires a locally nuanced understanding. However, when the questions are asked in local languages, we observe a decline in performance. Overall, there is significant room for improvement in this task.

\paragraph{Language Inference} Natural Language Inference (NLI) is a 3-class classification task, and typically smaller models perform close to random. The exception here is Cendol-7B, which performs subpar considering its size. We observe a similar trend to causal reasoning in terms of model performance across languages. Language Inference, particularly in local languages, remains a challenging task.

\paragraph{Reading Comprehension} This is perhaps one of the easier tasks, as we see models, especially the larger ones, achieving high performance. We also see minimal performance gaps between Indonesian and some other languages, including low-resource ones, where typically a larger drop is observed in the previously discussed tasks.

Our hypothesis is that this is an artifact of passage-based QA, where information can be retrieved correctly even if the question is not fully understood, for example by simply retrieving a person's name for a `who' question or a date for a `when' question. Nevertheless, we still observe some language gaps, and their performance leaves room for improvement, which highlights the usefulness of this task.

\paragraph{Open-Domain QA} We observe a noticeable performance drop across all models in Open-Domain QA, despite the questions being derived from the same set as those in reading comprehension. An interesting observation is the larger performance gap in Indonesian compared to other languages, suggesting that without any context, the model is unable to guess the answer. Unlike in reading comprehension, where the model can simply return dates or entities as plausible answers, Open-domain QA requires deeper reasoning.
% Trevor: is it open book QA or open domain?

\paragraph{Translation} In the translation task, we observe more comparable performance scores. Models are generally more similar in terms of performance, with the exception of smaller models such as Bloom 1B, Sailor 1B, and Qwen 1.5B and below. Some languages are noticeably more challenging, such as \texttt{bbc}, \texttt{bug}, \texttt{gor}, and \texttt{mak}.

\subsection{Result Across Languages}

%\begin{figure}[ht!]
%\centering
%\includegraphics[width=0.45\textwidth]{latex/figures/aggregate_Language.pdf}
%\caption{Aggregated performance across model}
%\label{fig:lang_result}
%\end{figure}

Focusing more on individual language performance across different tasks, we present the results across languages in Figure~\ref{fig:res_lang}.

Unlike the performance across models, we observe a more consistent performance trend across languages. if model performance is better on one language than another in one particular task, it tends to outperform across other tasks as well.

Models are visibly stronger in some languages. Javanese, one of the most resourced aside from Indonesia show strong performance. Betawi also performs well, as it closely resembles Indonesian; it is also commonly used as code-switching slang in everyday Indonesian, especially in social media.

Interestingly, however, performance does not always align with the number of speakers. Languages like Musi and Banjar perform quite strongly, despite having fewer speakers and being less explored in NLP research compared to more commonly studied languages like Javanese or Sundanese.

\subsection{Formal vs Informal Register}

\begin{figure}
\centering
\includegraphics[width=0.49\textwidth]{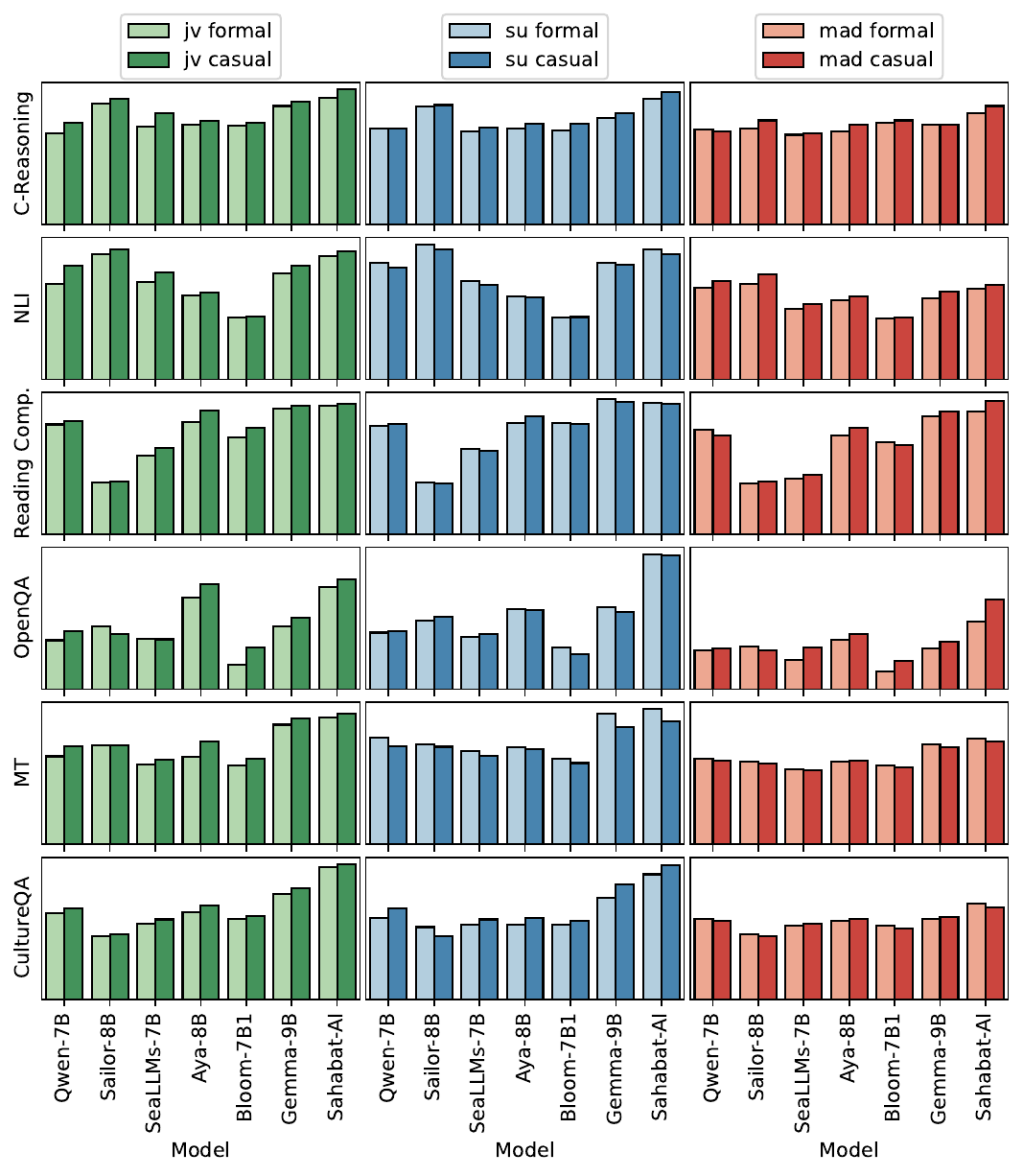}
\caption{Formal vs Casual register comparison; bars show the performance of different models.}
% Trevor: raw numbers should be in a table in the appendix
\label{fig:register_result}
\end{figure}

The results across different formality registers are shown in Figure~\ref{fig:register_result}. For Javanese, it is clear and consistent across all models and tasks that casual (Ngoko) Javanese is better handled. This finding aligns with~\citet{aji-etal-2022-one}, who found that the Ngoko register is easier to handle, specifically for language identification.
We argue that this is because Ngoko Javanese is the everyday variant commonly used in conversation and more readily available in data sources~\footnote{For example, the Javanese Wikipedia is mostly written in the casual register~\cite{farhansyah2025language}} In contrast, the formal variant is less commonly used in textual form and more situational.

However, we do not see a consistent pattern for Sundanese and Madurese. In contrast to Javanese, both Sundanese and Madurese speakers use their formal registers more often, including on the internet. Namely, while the use of formal Javanese might feel awkward in day-to-day settings, formal Sundanese and Madurese are more commonly used. Notably, the Sundanese Wikipedia is also written in a casual register. We hypothesize that, due to this, their performance is more situational.

\section{Conclusion}

We propose \datasetname, a novel benchmark for Indonesian low-resource languages. Our benchmark covers 20 local languages, three of which include two distinct politeness levels. We address six tasks: reading comprehension, open-domain question answering, natural language inference, cultural causal reasoning, cultural question answering, and machine translation. We evaluate a range of multilingual and region-specific LLMs, revealing substantial gaps and opportunities for improvement.

We hope that this benchmark will serve as a catalyst for future research and attention in low-resource NLP especially for Indonesian languages. By providing a comprehensive evaluation suite, we aim to encourage the community to build models that better capture the nuances of Indonesian local languages and cultures. In doing so, we envision \datasetname contributing to the broader goal of equitable language technology that benefits underrepresented communities globally.

\section{Limitations}

Our benchmark includes 20 Indonesian languages, which represent only a small portion of the 700+ languages spoken across the country. While not exhaustive, this selection aims to provide a starting point that reflects some linguistic and regional diversity. Additionally, the benchmark is currently limited to text data. We focus on this modality to ensure consistency and accessibility, while recognizing that future work could extend the dataset to include image, speech or other modalities. While our process may introduce some translationese, this risk is minimal given the use of expert translators working between closely related languages. Moreover, the parallel nature of the data enables comparison across languages. Our data is sourced from Indonesian-originated content, which should capture local nuances better than English-centric data. However, we acknowledge that it may not fully reflect the diverse cultural nuances of Indonesia, particularly those specific to each language.

\bibliography{anthology,custom}

\appendix
\onecolumn
\newpage

\section{Full Results}
\label{sec:appendix}

\begin{figure*}[ht!]
\centering
\includegraphics[width=0.99\textwidth]{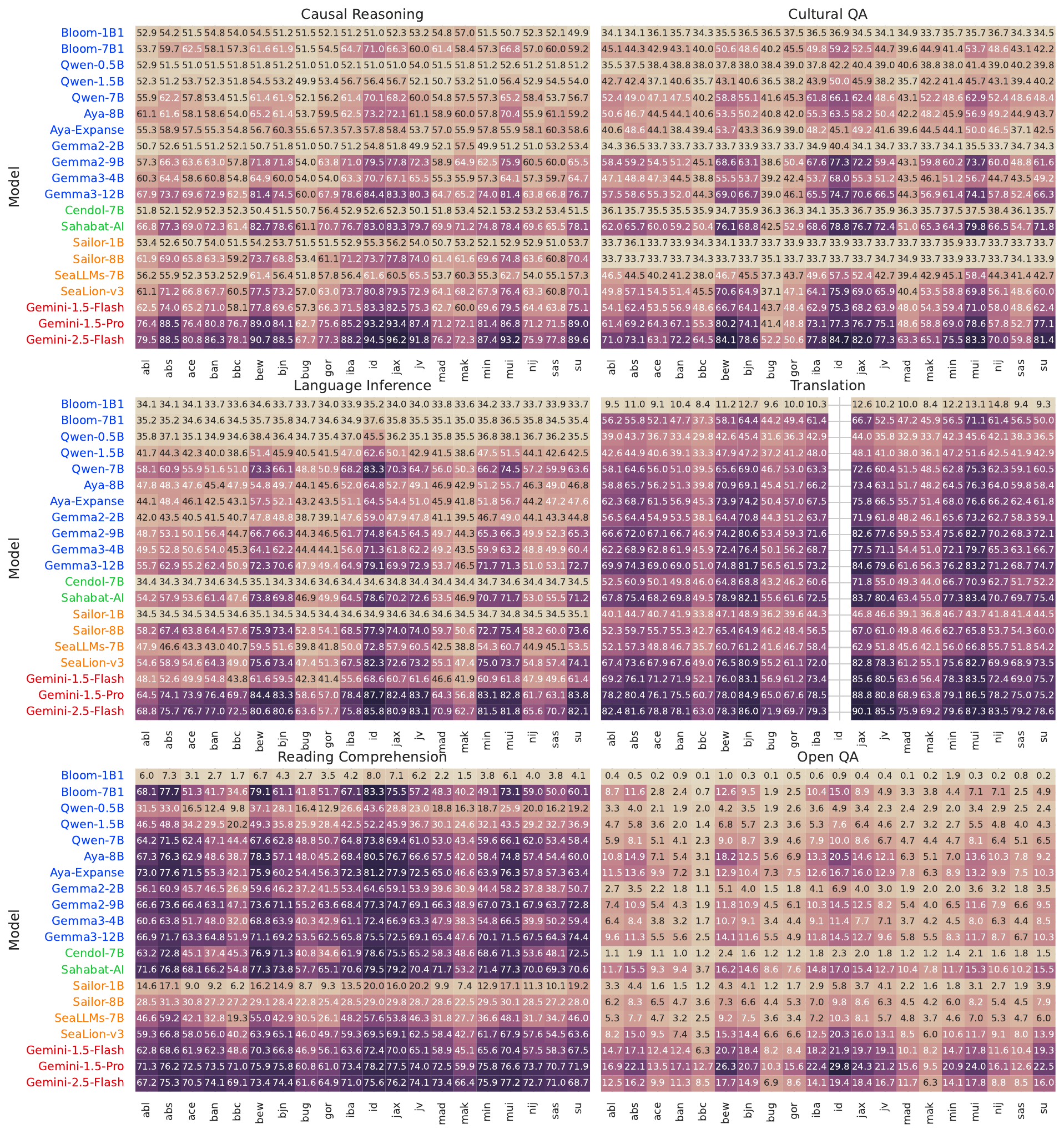}
\caption{Performance across all 6 tasks in \datasetname across different languages. We compare {\color{blue}{Multilingual models}}, {\color{green}{Indonesian-specific models}}, {\color{orange}{SEA-specific models}}, and {\color{red}{Commercial models}}}
\label{fig:main_result}
\end{figure*}

Table~\ref{fig:main_result} shows results across models and tasks using the best-performing prompts. It generally shows that Indonesian and some other languages are consistently easier for most models, whereas languages like Buginese (bug) are challenging. Gemini-1.5 Pro, Sahabat AI, and Gemma 9B show strong performance.

\newpage
\section{Model Configuration}
\label{appendix:models}

The following is the models used in this work. All benchmarking was done on a single A100, except for Gemini models in which we access via an API.

\begin{table}[ht!]
    \centering
    \begin{tabular}{ll}
        \toprule
        \textbf{Model} & \textbf{Hugging Face Checkpoint} \\
        \midrule
        Qwen\_500M\_instruct & \href{https://huggingface.co/Qwen/Qwen2.5-0.5B-Instruct}{Qwen/Qwen2.5-0.5B-Instruct} \\
        Qwen\_1\_5B\_instruct & \href{https://huggingface.co/Qwen/Qwen2.5-1.5B-Instruct}{Qwen/Qwen2.5-1.5B-Instruct} \\
        Qwen\_7B\_instruct & \href{https://huggingface.co/Qwen/Qwen2.5-7B-Instruct}{Qwen/Qwen2.5-7B-Instruct} \\
        Aya\_23\_8B & \href{https://huggingface.co/CohereForAI/aya-23-8B}{CohereForAI/aya-23-8B} \\
        Aya\_expanse\_8B & \href{https://huggingface.co/CohereForAI/aya-expanse-8B}{CohereForAI/aya-expanse-8B} \\
        BLOOMZ\_1B & \href{https://huggingface.co/bigscience/bloomz-1b1}{bigscience/bloomz-1b1} \\
        BLOOMZ\_7B & \href{https://huggingface.co/bigscience/bloomz-7b1}{bigscience/bloomz-7b1} \\
        gemma2\_2b & \href{https://huggingface.co/google/gemma-2-2b-it}{google/gemma-2-2b-it} \\
        gemma2\_9b & \href{https://huggingface.co/google/gemma-2-9b-it}{google/gemma-2-9b-it} \\
        Sailor2\_1B & \href{https://huggingface.co/sail/Sailor2-1B-Chat}{sail/Sailor2-1B-Chat} \\
        Sailor2\_8B & \href{https://huggingface.co/sail/Sailor2-8B-Chat}{sail/Sailor2-8B-Chat} \\
        SeaLLM\_v3\_7B & \href{https://huggingface.co/SeaLLMs/SeaLLMs-v3-7B-Chat}{SeaLLMs/SeaLLMs-v3-7B-Chat} \\
        Sea\_lion\_v3 & \href{https://huggingface.co/aisingapore/gemma2-9b-cpt-sea-lionv3-instruct}{aisingapore/gemma2-9b-cpt-sea-lionv3-instruct} \\
        Sahabat\_AI & \href{https://huggingface.co/GoToCompany/gemma2-9b-cpt-sahabatai-v1-instruct}{GoToCompany/gemma2-9b-cpt-sahabatai-v1-instruct} \\
        Cendol\_7B & \href{https://huggingface.co/indonlp/cendol-llama2-7b-inst}{indonlp/cendol-llama2-7b-inst} \\
        \bottomrule
    \end{tabular}
    \caption{Models and their corresponding Hugging Face checkpoints}
    \label{tab:models}
\end{table}

\section{Annotation}
\label{appendix:annotation}

Annotators are hired through a professional vendor, in which we pay them about \$0.8 per sentence translated and \$0.3 per sentence reviewed (prices in USD). Annotators are native in both Indonesian and the corresponding local languages (see Table~\ref{tab:demographics}). We hire 8-31 annotators per-language, with generally balanced gender distribution.
% Trevor: it said "city" rather than region, but the table below shows broader regions in the main (modulo Jakarta)

\iffalse
\begin{table}[]
\centering
\small
    \begin{tabular}{lcccc}
\hline

\textbf{Language} & \textbf{\#~Annotators} & \textbf{Demography} & \textbf{Male} & \textbf{Female} \\
\hline
Javanese (jv) Krama (Formal) & 16 & Jawa barat & 6 & 10 \\
Javanese (jv) Ngoko (Informal) & 24 & Jawa barat & 9 & 15 \\
Sundanese (su) Lemes (Formal) & 16 & Jawa & 10 & 6 \\
Sundanese (su) Loma (Informal) & 16 & Jawa & 10 & 6 \\
Banjar (bjn) & 16 & Kalimantan selatan & 7 & 9 \\
Madurese (mad) Enggi Enten (Formal) & 8 & Madura & 3 & 5 \\
Madurese (mad) Enja'Iya (Informal) & 16 & Surabaya & 9 & 7 \\
Minangkabau (min) & 24 & Jambi & 4 & 20 \\
Betawi (bew) & 24 & Jakarta & 5 & 19 \\
Buginese (bug) & 31 & Sulawesi selatan & 11 & 20 \\
Makasar (mak) & 24 & Sumatera & 6 & 18 \\
Acehnese (ace) & 24 & Jakarta & 8 & 16 \\
Balinese (ban) & 16 & Lombok & 9 & 7 \\
Musi (mui) & 16 & Jambi & 8 & 8 \\
Lampung Nyo (abl) & 16 & Lampung & 2 & 14 \\
Ambonese Malay (abs) & 16 & Jakarta & 4 & 12 \\
Batak Toba (bbc) & 24 & Jawa timur & 6 & 18 \\
Iban (iba) & 16 & Bali & 9 & 7 \\
Sasak (sas) & 16 & Lonbok & 3 & 13 \\
Gorontalo (gor) & 31 & Sumatra & 14 & 17 \\
Jambi Malay (jax) & 24 & Jambi & 8 & 16 \\
Ngaju (nij) & 24 & Jakarta & 7 & 17 \\
\hline
\end{tabular}
\caption{Annotator demographics for benchmark translation. Some of them do not live in the regions where the languages are mainly spoken, as migration within Indonesia is common.}
\label{tab:demographics}
\end{table}
\fi

\begin{table}[]
\centering
\small
    \begin{tabular}{lccc}
\hline
\textbf{Language} & \textbf{\#~Annotators} & \textbf{Male} & \textbf{Female} \\
\hline
Javanese (jv) Krama (Formal) & 16 & 6 & 10 \\
Javanese (jv) Ngoko (Informal) & 24 & 9 & 15 \\
Sundanese (su) Lemes (Formal) & 16 & 10 & 6 \\
Sundanese (su) Loma (Informal) & 16 & 10 & 6 \\
Banjar (bjn) & 16 & 7 & 9 \\
Madurese (mad) Enggi Enten (Formal) & 8 & 3 & 5 \\
Madurese (mad) Enja'Iya (Informal) & 16 & 9 & 7 \\
Minangkabau (min) & 24 & 4 & 20 \\
Betawi (bew) & 24 & 5 & 19 \\
Buginese (bug) & 31 & 11 & 20 \\
Makasar (mak) & 24 & 6 & 18 \\
Acehnese (ace) & 24 & 8 & 16 \\
Balinese (ban) & 16 & 9 & 7 \\
Musi (mui) & 16 & 8 & 8 \\
Lampung Nyo (abl) & 16 & 2 & 14 \\
Ambonese Malay (abs) & 16 & 4 & 12 \\
Batak Toba (bbc) & 24 & 6 & 18 \\
Iban (iba) & 16 & 9 & 7 \\
Sasak (sas) & 16 & 3 & 13 \\
Gorontalo (gor) & 31 & 14 & 17 \\
Jambi Malay (jax) & 24 & 8 & 16 \\
Ngaju (nij) & 24 & 7 & 17 \\
\hline
\end{tabular}
\caption{Annotator demographics for benchmark translation. Some of them do not live in the regions where the languages are mainly spoken, as migration within Indonesia is common.}
\label{tab:demographics}
\end{table}

Annotation is done through Google Sheet. In that sheet, we also implement script-based validation that will automatically detect potential inconsistencies for further discussion with annotators. We also put the overview guidance on the sheet as follow:

Please translate the text in the corresponding cell. Ensure that the meaning and semantics are preserved. Do not add to or remove any context from the text.

Suggested guidelines for formal vs informal registers
\begin{itemize}
    \item Translations into Javanese, Sundanese and Madurese include a formal vs informal register.
By this, we mean language used in every day/casual conversations, and that used in more formal/polite setting and in many written settings.
Generally, the casual one is the register people use to talk to their friends, and the formal one is used to talk to parents, boss, teacher, or strangers.
More specifically:
    \begin{itemize}
        \item Javanese: Ngoko for casual, Krama for formal
        \item Sundanese: Loma for casual, Lemes for formal
        \item Madurese: Enja'Iya for casual, and Enggi Enten for formal
    \end{itemize}
\end{itemize}

General guidelines
\begin{itemize}
    \item The translations should not include "/" in the translation to specify multiple translation options, instead choose one option (the most natural.)
    \item The translations should not include clarifications in brackets, or redundant information, alternative translations etc.
    \item 
Numbers should be translated without brackets in a format that matches the original Indonesian input (e.g., words vs numerals).
\end{itemize}

\section{Open QA Performance: Exact Match vs LLM-as-a-Judge}
\label{appendix:openqa}

For Open QA, we consider an answer correct if the gold label is a substring of the generated output. However, this approach may still be too strict. Therefore, we also analyze performance using an LLM-as-a-Judge evaluation. 
In this setting, we use Gemini-2.5-Flash as our judge. Specifically, we employ the following prompt:

\begin{verbatim}
You are an AI evaluator. Your task is to score a model's response for a factual question.
You will be given the question, a gold-standard answer, and the model's response.

Compare the model's response to the gold-standard answer.
Based on factual correctness, give a score from 1 to 5.

- 5: The answer is completely correct and aligns with the gold-standard.
- 4: The answer is almost correct with very minor inaccuracies.
- 3: The answer is partially correct but has noticeable errors.
- 2: The answer is mostly incorrect.
- 1: The answer is completely incorrect or irrelevant.

Respond with a single JSON object containing one key: "score". Do not add any other text.

---
**[EVALUATION TASK]**
**## Question:**
{question}
**## Gold-Standard Answer:**
{gold_standard}
**## Model's Response:**
{model_response}
\end{verbatim}

\begin{figure}
\centering
\includegraphics[width=0.69\textwidth]{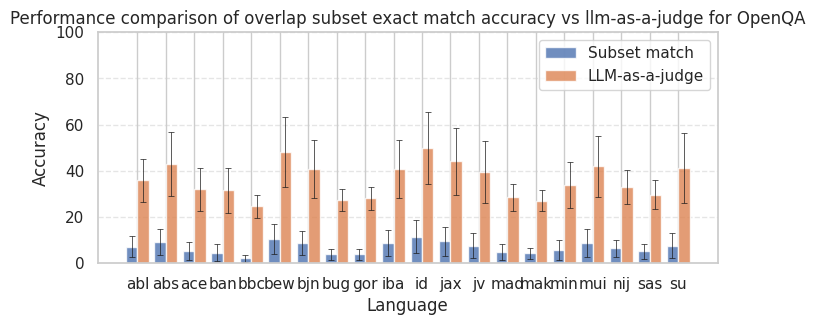}
\caption{Comparing performance of subset match vs LLM-as-a-judge across different languages}
\label{fig:llm_judge1}
\end{figure}

\begin{figure}
\centering
\includegraphics[width=0.69\textwidth]{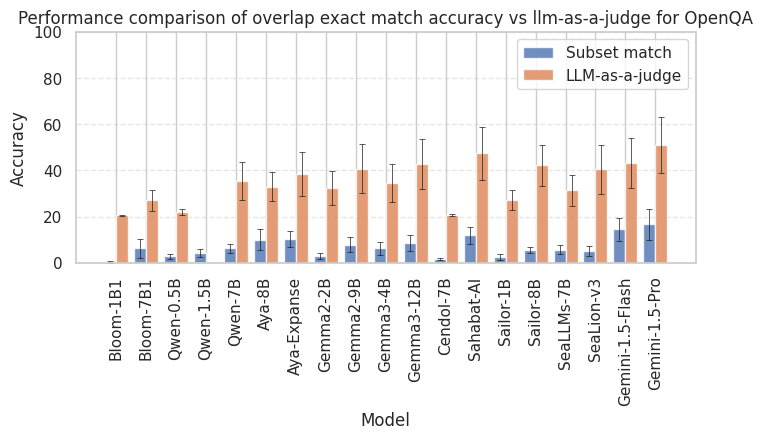}
\caption{Comparing performance of subset match vs LLM-as-a-judge across different models}
\label{fig:llm_judge2}
\end{figure}

The prompt returns a correctness score ranging from 1 to 5, which we normalize by dividing the score by 5. We observe a similar trend to the string subset approach, as shown in Figures~\ref{fig:llm_judge1} and~\ref{fig:llm_judge2}. However, an important consideration is whether LLM-as-a-Judge is a reliable evaluator for low-resource languages.

\section{Result Across Prompts}

\begin{figure}[ht!]
\centering
\includegraphics[width=0.49\textwidth]{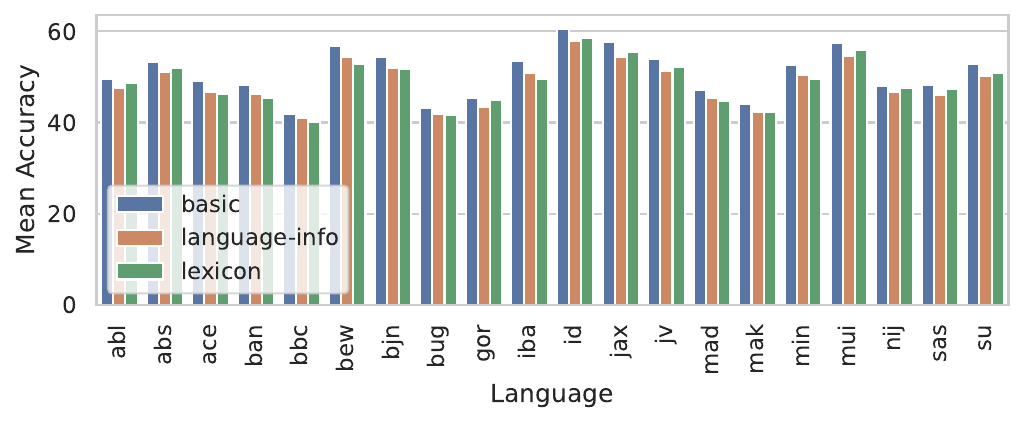}
\caption{Prompt variation performance across all model and tasks}
\label{fig:prompt_result}

\includegraphics[width=0.49\textwidth]{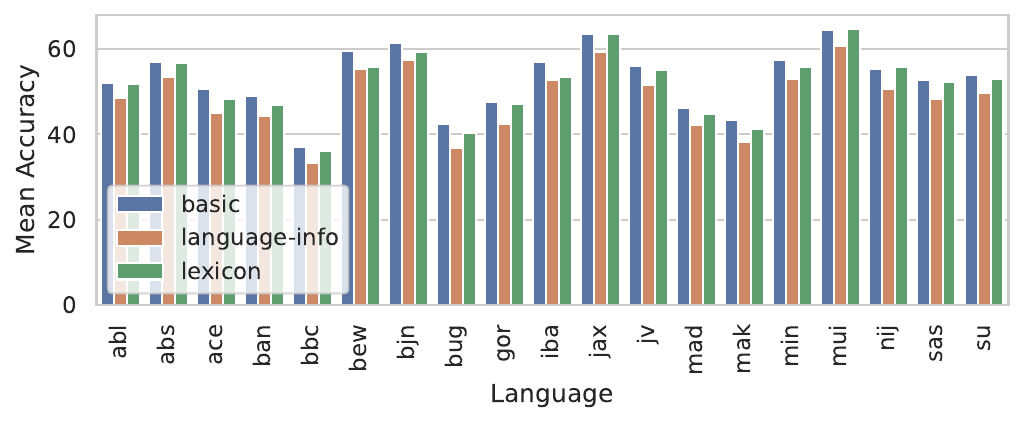}
\caption{Prompt variation performance across all model for MT task}
\label{fig:prompt_result-mt}
\end{figure}

We explored various prompt strategies to improve performance. Interestingly, explicitly informing the model about the language situation did not lead to a meaningful improvement, and instead we see a degradation, as shown in Figure~\ref{fig:prompt_result-mt}. This held true whether we provided direct information or included lexicon information.

We see a consistent degradation across languages and models. However, we do note an interesting finding: lexicon information does not degrade MT performance as much. We hypothesize that the language-info prompt might be misleading. While it is true that local Indonesian languages often share vocabulary, the same can be said for the many false friends they share, which can confuse the model. Lexicon guidance might also confuse the model, as some tokens have multiple word translations depending on the word sense.

\begin{figure}[ht!]
\centering
\includegraphics[width=0.6\textwidth]{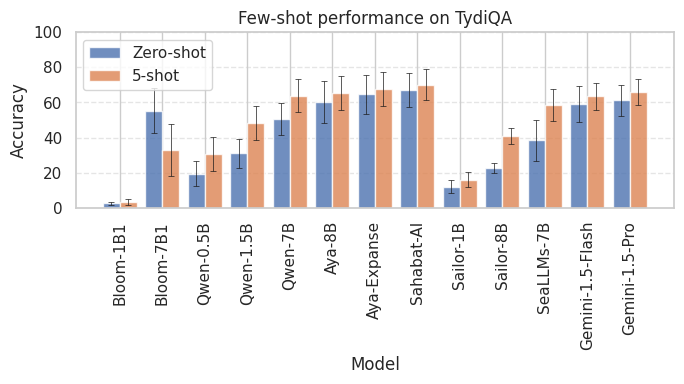}
\caption{Few-shot performance on reading comprehension}
\label{fig:few-shot}
\end{figure}

Perhaps unexpectedly, few-shot prompting yields noticeable performance gains (Figure~\ref{fig:few-shot}), although this experiment was conducted on a limited set of reading comprehension examples due to resource constraints. These improvements are consistent across models.

\section{Result Across Registers}
\label{sec:register_full}

The following Table~\ref{tab:model_performance1} to Table~\ref{tab:model_performance6} show models' performance across formal vs casual registers on Javanese, Madurese, and Sundanese.

\begin{table}
\small
\center
\begin{tabular}{lcccccc}
\toprule
% Model & jv casual  & jv formal & mad casual  & mad formal  & su casual  & su formal  \\
Model & \multicolumn{2}{c}{jav} & \multicolumn{2}{c}{jav} & \multicolumn{2}{c}{jav} \\
& I & F & I & F & I & F \\
\midrule
Aya-8B & 61.10 & 58.90 & 58.90 & 54.79 & 59.18 & 56.44 \\
Aya-Expanse & 53.70 & 53.97 & 56.99 & 54.52 & 58.63 & 55.62 \\
Bloom-1B1 & 53.15 & 55.34 & 54.79 & 52.88 & 49.86 & 50.68 \\
Bloom-7B1 & 60.00 & 58.36 & 61.37 & 59.73 & 59.18 & 55.62 \\
Cendol-7B & 50.14 & 52.05 & 51.78 & 50.96 & 51.51 & 53.15 \\
Gemini-1.5-Flash & 75.34 & 72.33 & 62.74 & 58.36 & 75.07 & 69.32 \\
Gemini-1.5-Pro & 87.40 & 86.85 & 71.23 & 68.77 & 89.04 & 89.32 \\
Gemini-2.5-Flash & 91.78 & 90.41 & 76.16 & 73.15 & 89.59 & 88.49 \\
Gemini-M & 78.08 & 77.81 & 61.10 & 55.89 & 78.36 & 76.99 \\
Gemma2-2B & 49.86 & 50.41 & 52.05 & 49.04 & 53.42 & 49.59 \\
Gemma2-9B & 72.33 & 69.86 & 58.90 & 58.63 & 65.48 & 62.74 \\
Gemma3-12B & 80.27 & 79.18 & 64.66 & 61.92 & 76.71 & 76.16 \\
Gemma3-4B & 65.48 & 62.19 & 55.34 & 53.42 & 64.66 & 61.92 \\
Qwen-0.5B & 53.97 & 50.96 & 51.51 & 51.51 & 51.23 & 51.78 \\
Qwen-1.5B & 52.05 & 54.52 & 50.68 & 49.59 & 53.97 & 50.68 \\
Qwen-7B & 60.00 & 53.97 & 54.79 & 55.89 & 56.71 & 56.71 \\
Sahabat-AI & 79.73 & 74.79 & 69.86 & 65.48 & 78.08 & 74.25 \\
Sailor-1B & 53.97 & 55.89 & 50.68 & 51.78 & 53.70 & 52.60 \\
Sailor-8B & 73.97 & 71.23 & 61.37 & 56.71 & 70.41 & 69.59 \\
SeaLLMs-7B & 65.48 & 57.53 & 53.70 & 52.88 & 57.26 & 54.79 \\
SeaLion-v3 & 72.88 & 71.51 & 64.11 & 60.82 & 70.14 & 69.32 \\
\bottomrule
\end{tabular}

\caption{Model Performance on Causal Reasoning by Language Style}
\label{tab:model_performance1}
\end{table}

\begin{table}
\small
\center
\begin{tabular}{lcccccc}
\toprule
% Model & jv casual  & jv formal & mad casual  & mad formal  & su casual  & su formal  \\
Model & \multicolumn{2}{c}{jav} & \multicolumn{2}{c}{jav} & \multicolumn{2}{c}{jav} \\
& I & F & I & F & I & F \\
\midrule
Aya-8B & 49.10 & 47.65 & 46.89 & 44.88 & 46.82 & 47.16 \\
Aya-Expanse & 50.97 & 45.78 & 45.92 & 42.12 & 47.65 & 48.13 \\
Bloom-1B1 & 33.96 & 33.61 & 33.82 & 33.96 & 33.68 & 33.96 \\
Bloom-7B1 & 35.55 & 34.92 & 35.06 & 34.85 & 35.41 & 35.13 \\
Cendol-7B & 34.44 & 34.44 & 34.44 & 34.44 & 34.51 & 34.51 \\
Gemini-1.5-Flash & 61.62 & 58.64 & 46.61 & 44.81 & 61.41 & 63.07 \\
Gemini-1.5-Pro & 83.68 & 81.88 & 64.32 & 59.75 & 83.82 & 84.16 \\
Gemini-2.5-Flash & 83.13 & 81.33 & 70.89 & 71.16 & 82.09 & 82.43 \\
Gemini-M & 73.93 & 69.71 & 55.81 & 52.14 & 72.68 & 74.41 \\
Gemma2-2B & 47.79 & 42.19 & 41.08 & 39.00 & 44.81 & 44.26 \\
Gemma2-9B & 64.52 & 60.37 & 49.65 & 46.13 & 65.28 & 66.11 \\
Gemma3-12B & 72.89 & 68.05 & 53.67 & 47.79 & 72.68 & 74.76 \\
Gemma3-4B & 62.17 & 56.92 & 49.17 & 45.23 & 60.44 & 63.42 \\
Qwen-0.5B & 35.06 & 35.06 & 35.82 & 34.99 & 35.55 & 36.17 \\
Qwen-1.5B & 42.95 & 40.53 & 41.49 & 41.29 & 42.46 & 42.88 \\
Qwen-7B & 64.66 & 54.15 & 56.02 & 52.01 & 63.55 & 66.04 \\
Sahabat-AI & 72.61 & 70.12 & 53.53 & 51.45 & 71.23 & 73.79 \\
Sailor-1B & 34.58 & 34.79 & 34.58 & 34.58 & 35.13 & 34.92 \\
Sailor-8B & 74.00 & 71.30 & 59.68 & 54.01 & 73.58 & 76.76 \\
SeaLLMs-7B & 60.51 & 55.12 & 42.53 & 40.18 & 53.46 & 56.09 \\
SeaLion-v3 & 73.17 & 67.57 & 55.12 & 49.86 & 74.07 & 74.62 \\
\bottomrule
\end{tabular}

\caption{Model Performance on NLI by Language Style}
\label{tab:model_performance2}
\end{table}

\begin{table}
\small
\center
\begin{tabular}{lcccccc}
\toprule
% Model & jv casual  & jv formal & mad casual  & mad formal  & su casual  & su formal  \\
Model & \multicolumn{2}{c}{jav} & \multicolumn{2}{c}{jav} & \multicolumn{2}{c}{jav} \\
& I & F & I & F & I & F \\
\midrule
Aya-8B & 69.68 & 67.18 & 60.32 & 58.55 & 62.79 & 67.55 \\
Aya-Expanse & 75.10 & 68.76 & 68.84 & 67.26 & 63.36 & 67.28 \\
Bloom-1B1 & 6.18 & 3.62 & 2.32 & 2.98 & 4.08 & 4.36 \\
Bloom-7B1 & 57.20 & 52.14 & 48.26 & 49.42 & 60.10 & 59.50 \\
Cendol-7B & 65.24 & 61.93 & 58.26 & 52.91 & 72.53 & 71.00 \\
Gemini-1.5-Flash & 69.76 & 69.12 & 61.22 & 60.04 & 70.18 & 71.45 \\
Gemini-1.5-Pro & 74.00 & 73.85 & 72.52 & 69.51 & 71.91 & 74.75 \\
Gemini-2.5-Flash & 74.08 & 74.18 & 73.39 & 73.89 & 68.69 & 68.28 \\
Gemini-M & 72.70 & 74.51 & 66.21 & 64.61 & 74.85 & 73.92 \\
Gemma2-2B & 58.82 & 53.56 & 48.48 & 47.01 & 56.30 & 56.60 \\
Gemma2-9B & 69.08 & 67.53 & 66.35 & 63.52 & 72.76 & 71.32 \\
Gemma3-12B & 69.06 & 72.00 & 65.40 & 60.15 & 74.41 & 73.54 \\
Gemma3-4B & 63.26 & 57.46 & 47.93 & 47.17 & 59.40 & 59.17 \\
Qwen-0.5B & 31.05 & 23.96 & 22.32 & 21.09 & 25.84 & 28.76 \\
Qwen-1.5B & 50.05 & 44.04 & 46.99 & 40.87 & 45.36 & 48.41 \\
Qwen-7B & 62.43 & 59.16 & 59.91 & 60.76 & 70.27 & 68.51 \\
Sahabat-AI & 76.50 & 73.36 & 71.74 & 70.72 & 75.37 & 75.19 \\
Sailor-1B & 20.21 & 17.61 & 13.16 & 11.18 & 19.91 & 19.87 \\
Sailor-8B & 45.41 & 45.85 & 44.05 & 42.31 & 45.49 & 44.45 \\
SeaLLMs-7B & 67.22 & 58.09 & 54.63 & 54.07 & 59.88 & 63.15 \\
SeaLion-v3 & 62.52 & 63.04 & 58.36 & 53.12 & 63.57 & 62.40 \\
\bottomrule
\end{tabular}

\caption{Model Performance on TydiQA by Language Style}
\label{tab:model_performance3}
\end{table}

\begin{table}
\small
\center
\begin{tabular}{lcccccc}
\toprule
% Model & jv casual  & jv formal & mad casual  & mad formal  & su casual  & su formal  \\
Model & \multicolumn{2}{c}{jav} & \multicolumn{2}{c}{jav} & \multicolumn{2}{c}{jav} \\
& I & F & I & F & I & F \\
\midrule
Aya-8B & 13.05 & 12.02 & 9.88 & 7.98 & 9.11 & 9.69 \\
Aya-Expanse & 12.76 & 10.00 & 8.35 & 8.45 & 10.22 & 10.27 \\
Bloom-1B1 & 0.40 & 0.38 & 0.13 & 0.54 & 0.24 & 0.28 \\
Bloom-7B1 & 4.80 & 2.85 & 3.23 & 2.00 & 4.81 & 4.00 \\
Cendol-7B & 1.81 & 1.74 & 1.20 & 1.16 & 1.49 & 1.55 \\
Gemini-1.5-Flash & 26.76 & 26.20 & 20.43 & 19.57 & 26.46 & 24.07 \\
Gemini-1.5-Pro & 21.00 & 18.43 & 15.40 & 13.78 & 22.22 & 19.44 \\
Gemini-2.5-Flash & 16.52 & 15.48 & 11.61 & 10.30 & 15.81 & 15.07 \\
Gemini-M & 2.38 & 4.57 & 5.30 & 4.27 & 2.16 & 1.54 \\
Gemma2-2B & 3.01 & 2.00 & 1.93 & 1.42 & 3.51 & 3.92 \\
Gemma2-9B & 8.14 & 7.12 & 5.39 & 4.68 & 9.37 & 8.77 \\
Gemma3-12B & 9.51 & 8.77 & 5.69 & 4.31 & 10.24 & 8.77 \\
Gemma3-4B & 6.98 & 5.91 & 3.68 & 3.28 & 8.43 & 7.30 \\
Qwen-0.5B & 2.28 & 3.04 & 2.89 & 2.03 & 2.40 & 2.90 \\
Qwen-1.5B & 5.92 & 5.29 & 3.93 & 4.10 & 4.62 & 4.40 \\
Qwen-7B & 7.24 & 7.25 & 7.73 & 7.94 & 8.57 & 8.36 \\
Sahabat-AI & 23.33 & 22.65 & 20.34 & 16.03 & 22.64 & 21.39 \\
Sailor-1B & 4.10 & 3.06 & 2.51 & 2.03 & 3.86 & 3.98 \\
Sailor-8B & 8.83 & 8.09 & 7.13 & 6.20 & 8.21 & 8.31 \\
SeaLLMs-7B & 8.46 & 6.88 & 7.33 & 5.10 & 9.89 & 9.33 \\
SeaLion-v3 & 23.42 & 20.26 & 18.22 & 15.47 & 22.13 & 21.79 \\
\bottomrule
\end{tabular}

\caption{Model Performance on Open QA by Language Style}
\label{tab:model_performance4}
\end{table}

\begin{table}
\small
\center
\begin{tabular}{lcccccc}
\toprule
% Model & jv casual  & jv formal & mad casual  & mad formal  & su casual  & su formal  \\
Model & \multicolumn{2}{c}{jav} & \multicolumn{2}{c}{jav} & \multicolumn{2}{c}{jav} \\
& I & F & I & F & I & F \\
\midrule
Aya-8B & 63.12 & 53.76 & 51.70 & 51.06 & 58.43 & 59.56 \\
Aya-Expanse & 66.53 & 59.33 & 55.65 & 55.89 & 61.80 & 65.00 \\
Bloom-1B1 & 10.21 & 8.64 & 10.02 & 10.92 & 9.26 & 10.91 \\
Bloom-7B1 & 52.47 & 48.74 & 47.22 & 48.77 & 49.97 & 52.86 \\
Cendol-7B & 54.98 & 53.48 & 49.28 & 47.41 & 52.20 & 46.26 \\
Gemini-1.5-Flash & 80.53 & 79.30 & 63.59 & 67.24 & 75.74 & 83.76 \\
Gemini-1.5-Pro & 80.79 & 84.47 & 68.89 & 76.39 & 75.20 & 86.86 \\
Gemini-2.5-Flash & 85.52 & 85.95 & 75.94 & 81.91 & 78.62 & 88.18 \\
Gemma2-2B & 61.82 & 55.79 & 48.17 & 49.49 & 59.06 & 64.10 \\
Gemma2-9B & 77.61 & 73.57 & 59.47 & 61.62 & 72.15 & 80.49 \\
Gemma3-12B & 79.56 & 77.30 & 61.64 & 64.32 & 74.71 & 83.56 \\
Gemma3-4B & 71.15 & 68.42 & 54.41 & 55.90 & 66.67 & 75.99 \\
Qwen-0.5B & 35.81 & 35.25 & 32.90 & 33.69 & 36.54 & 38.54 \\
Qwen-1.5B & 41.00 & 38.12 & 38.04 & 38.34 & 42.92 & 41.31 \\
Qwen-7B & 60.43 & 54.14 & 51.48 & 52.51 & 60.49 & 65.46 \\
Sahabat-AI & 80.44 & 77.76 & 63.42 & 64.94 & 75.39 & 83.18 \\
Sailor-1B & 46.57 & 44.77 & 39.08 & 39.38 & 44.53 & 47.45 \\
Sailor-8B & 61.02 & 60.77 & 49.76 & 51.01 & 60.04 & 61.76 \\
SeaLLMs-7B & 51.80 & 49.21 & 45.56 & 46.39 & 54.21 & 57.55 \\
SeaLion-v3 & 78.27 & 73.94 & 61.20 & 62.30 & 73.48 & 81.01 \\
\bottomrule
\end{tabular}

\caption{Model Performance on Machine Translation by Language Style}
\label{tab:model_performance5}
\end{table}

\begin{table}
\small
\center
\begin{tabular}{lcccccc}
\toprule
% Model & jv casual  & jv formal & mad casual  & mad formal  & su casual  & su formal  \\
Model & \multicolumn{2}{c}{jav} & \multicolumn{2}{c}{jav} & \multicolumn{2}{c}{jav} \\
& I & F & I & F & I & F \\
\midrule
Aya-8B & 50.39 & 46.47 & 42.16 & 42.94 & 43.73 & 40.00 \\
Aya-Expanse & 41.57 & 41.76 & 39.61 & 39.02 & 42.55 & 40.78 \\
Bloom-1B1 & 34.12 & 37.45 & 34.90 & 37.25 & 34.51 & 37.25 \\
Bloom-7B1 & 44.71 & 43.14 & 39.61 & 37.84 & 42.16 & 40.00 \\
Cendol-7B & 35.88 & 34.71 & 36.27 & 36.47 & 35.69 & 36.86 \\
Gemini-1.5-Flash & 63.92 & 61.76 & 48.04 & 44.31 & 62.35 & 64.12 \\
Gemini-1.5-Pro & 75.10 & 67.65 & 48.63 & 52.16 & 77.06 & 71.57 \\
Gemini-2.5-Flash & 77.25 & 78.43 & 63.33 & 65.88 & 81.37 & 78.82 \\
Gemma2-2B & 34.71 & 33.73 & 33.73 & 33.73 & 34.31 & 34.71 \\
Gemma2-9B & 59.41 & 56.27 & 43.14 & 43.92 & 61.57 & 54.12 \\
Gemma3-12B & 66.47 & 60.98 & 44.31 & 41.76 & 66.27 & 60.39 \\
Gemma3-4B & 51.18 & 49.41 & 43.53 & 39.41 & 49.22 & 44.51 \\
Qwen-0.5B & 39.02 & 39.80 & 40.59 & 39.80 & 39.80 & 36.27 \\
Qwen-1.5B & 38.24 & 37.45 & 35.69 & 36.67 & 40.20 & 39.22 \\
Qwen-7B & 48.63 & 46.08 & 43.14 & 41.96 & 48.43 & 43.73 \\
Sahabat-AI & 72.35 & 70.78 & 50.98 & 49.22 & 71.76 & 66.86 \\
Sailor-1B & 33.73 & 33.73 & 33.73 & 33.73 & 33.73 & 33.73 \\
Sailor-8B & 34.71 & 33.73 & 34.90 & 33.73 & 33.92 & 38.63 \\
SeaLLMs-7B & 42.75 & 40.59 & 39.41 & 40.39 & 42.75 & 40.00 \\
SeaLion-v3 & 65.88 & 59.80 & 40.39 & 42.16 & 60.00 & 54.51 \\
\bottomrule
\end{tabular}

\caption{Model Performance on Cultural QA by Language Style}
\label{tab:model_performance6}
\end{table}

\section{Prompt Configuration}

\subsection*{1. Sentence Completion Prompts}

\subsubsection*{Variant: lexicon}
\begin{verbatim}
How would you continue the {language} sentence "{context}"?
{lexicon_hint}

Choice A: {choice1}
Choice B: {choice2}
Choice C: {choice3}

Answer with either A, B, or C:
\end{verbatim}

\subsubsection*{Variant: basic}
\begin{verbatim}
How would you continue the {language} sentence "{context}"?

Choice A: {choice1}
Choice B: {choice2}
Choice C: {choice3}

Answer with either A, B, or C:
\end{verbatim}

\subsubsection*{Variant: language-info}
\begin{verbatim}
Determine the follow-up sentence in {language}.
It is similar to Indonesian. They share similar grammar and vocabulary. 
If you encounter unfamiliar words, consider their Indonesian equivalents.

How would you continue the {language} sentence "{context}"?

Choice A: {choice1}
Choice B: {choice2}
Choice C: {choice3}

Answer with either A, B, or C:
\end{verbatim}

\subsubsection*{Variant: cloze}
\begin{verbatim}
{context}
\end{verbatim}

\subsection*{2. Causal Reasoning Prompts}

\subsubsection*{Variant: lexicon}
\begin{verbatim}
Determine the cause/effect of a given premise in {language}.
{lexicon_hint}

Premise: {premise}

Choice A: {choice1}
Choice B: {choice2}

Question: Which is the more likely {question_type}, given the premise? 
Answer with either A or B:
\end{verbatim}

\subsubsection*{Variant: basic}
\begin{verbatim}
Premise: {premise}

Choice A: {choice1}
Choice B: {choice2}

Question: Which is the more likely {question_type}, given the premise? 
Answer with either A or B:
\end{verbatim}

\subsubsection*{Variant: language-info}
\begin{verbatim}
Determine the cause/effect of a given premise in {language}.
It is similar to Indonesian. They share similar grammar and vocabulary. 
If you encounter unfamiliar words, consider their Indonesian equivalents.

Premise: {premise}

Choice A: {choice1}
Choice B: {choice2}

Question: Which is the more likely {question_type}, given the premise? 
Answer with either A or B:
\end{verbatim}

\subsubsection*{Variant: cloze}
\begin{verbatim}
{premise}, {question_type_verbose}
\end{verbatim}

\subsection*{3. Translation Prompts}

\subsubsection*{Variant: lexicon}
\begin{verbatim}
Translate the following {language} text into Indonesian. 
Please translate the input directly without any other comments.
{lexicon_hint}

Input: {source}
Output:
\end{verbatim}

\subsubsection*{Variant: basic}
\begin{verbatim}
Translate the following {language} text into Indonesian. 
Please translate the input directly without any other comments.

Input: {source}
Output:
\end{verbatim}

\subsubsection*{Variant: language-info}
\begin{verbatim}
Translate the following {language} text into Indonesian. 
Please translate the input directly without any other comments.
They share similar grammar and vocabulary. If you encounter unfamiliar 
words, consider just copying the original words.

Input: {source}
Output:
\end{verbatim}

\subsection*{4. NLI Prompts}

\subsubsection*{Variant: lexicon}
\begin{verbatim}
Determine the relationship between the following two statements 
written in {language}.
{lexicon_hint}

The relationship can be one of the following:

- Entailment: The premise implies the hypothesis.
- Contradiction: The premise contradicts the hypothesis.
- Neutral: The premise and hypothesis are not related or the relationship 
  cannot be concluded.

Premise: {premise}
Hypothesis: {hypothesis}

Relationship (Please answer concisely with Entailment, Contradiction, or Neutral):
\end{verbatim}

\subsubsection*{Variant: basic}
\begin{verbatim}
Determine the relationship between the following two statements.
The relationship can be one of the following:

- Entailment: The premise implies the hypothesis.
- Contradiction: The premise contradicts the hypothesis.
- Neutral: The premise and hypothesis are not related or the relationship 
  cannot be concluded.

Premise: {premise}
Hypothesis: {hypothesis}

Relationship (Please answer concisely with Entailment, Contradiction, or Neutral):
\end{verbatim}

\subsubsection*{Variant: language-info}
\begin{verbatim}
Determine the relationship between the following two statements 
written in {language}.
It is similar to Indonesian. They share similar grammar and vocabulary. 
If you encounter unfamiliar words, consider their Indonesian equivalents.

The relationship can be one of the following:

- Entailment: The premise implies the hypothesis.
- Contradiction: The premise contradicts the hypothesis.
- Neutral: The premise and hypothesis are not related or the relationship 
  cannot be concluded.

Premise: {premise}
Hypothesis: {hypothesis}

Relationship (Please answer concisely with Entailment, Contradiction, or Neutral):
\end{verbatim}

\subsection*{5. QA Prompts}

\subsubsection*{Variant: lexicon}
\begin{verbatim}
Answer the following question concisely.

The question is in {language}.

The following lexicon might help you understand the language better:

{lexicon_hint}

Question: {question}
\end{verbatim}

\subsubsection*{Variant: basic}
\begin{verbatim}
Answer the following question concisely.

{question}
\end{verbatim}

\subsubsection*{Variant: language-info}
\begin{verbatim}
Answer the following question concisely.

The question is in {language}, which is similar to Indonesian. 
They share similar grammar and vocabulary. If you encounter 
unfamiliar words, consider their Indonesian equivalents.

Question: {question}
\end{verbatim}

\subsection*{6. QA Extraction Prompts}

\subsubsection*{Variant: lexicon}
\begin{verbatim}
Extract the answer to the following question from the given context.

The question is in {language}.
{lexicon_hint}

Context:
{context}

Now, extract the answer to the following question from the given context.
Be concise and straightforward; no explanations are needed.

Question:
{question}
\end{verbatim}

\subsubsection*{Variant: basic}
\begin{verbatim}
Context:
{context}

Extract the answer to the following question from the given context.
Be concise and straightforward; no explanations are needed.

Question:
{question}
\end{verbatim}

\subsubsection*{Variant: language-info}
\begin{verbatim}
Extract the answer to the following question from the given context.

The question is in {language}, which is similar to Indonesian. 
They share similar grammar and vocabulary. If you encounter 
unfamiliar words, consider their Indonesian equivalents.

Context:
{context}

Now, extract the answer to the following question from the given context.
Be concise and straightforward; no explanations are needed.

Question:
{question}
\end{verbatim}

\end{document}